\documentclass[final,5p,times,twocolumn]{elsarticle}

\usepackage{amssymb}
\usepackage{bm}
\usepackage{ bbold }
\usepackage{multirow}
\usepackage{caption}
\usepackage{subfigure}
\usepackage{booktabs}
\usepackage{float}
\usepackage{color}
\usepackage[table,xcdraw]{xcolor}
\usepackage{booktabs}

\journal{Neurocomputing}

\begin{document}

\begin{frontmatter}



\author[1]{Pengfei Ma}
\address[1]{School of Artificial Intelligence, Hebei University of   Technology,
	Tianjing,
	300401, 
	China}

\author[1,2]{Youxi Wu \corref{mycorrespondingauthor}}

\cortext[mycorrespondingauthor]{Corresponding author}
\ead{wuc567@163.com}
\address[2]{Hebei Key Laboratory of Big Data Computing,
	Tianjing,
	300401, 
	China}

\author[3]{Yan Li}
\address[3]{School of Economics and Management, Hebei University of Technology,
	Tianjing,
	300401, 
	China}

\author[4]{Lei Guo}
\address[4]{State Key Laboratory of Reliability and Intelligence of Electrical Equipment, Hebei University of Technology,
	Tianjing,
	300401, 
	China}

\author[5]{Zhao Li}
\address[5]{Alibaba Group,
	Hangzhou,
	310000, 
	Zhejiang Province,
	China}

\title{DBC-Forest: Deep forest with binning confidence screening }

\begin{abstract}
	As a deep learning model, deep confidence screening forest (gcForestcs) has achieved great success in various applications. Compared with the traditional deep forest approach, gcForestcs effectively reduces the high time cost by passing some instances in the high-confidence region directly to the final stage. However, there is a group of instances with low accuracy in the high-confidence region, which are called mis-partitioned instances. To find these mis-partitioned instances, this paper proposes a deep binning confidence screening forest (DBC-Forest) model, which packs all instances into bins based on their confidences. In this way, more accurate instances can be passed to the final stage, and the performance is improved. Experimental results show that DBC-Forest achieves highly accurate predictions for the same hyperparameters and is faster than other similar models to achieve the same accuracy.
\end{abstract}




\end{frontmatter}

\section{Introduction}
As an important field of artificial intelligence, deep learning has become a topic of research interest in various domains \cite{ref1,ref2,ref3}. Deep neural networks (DNNs) \cite{ref4} has better performance than traditional learning models \cite{ref5,tkdd2021,ins2021}, and rely on two main concepts: firstly, representation learning via layer-by-layer processing is used to extract effective features, and secondly, in-model feature transformation such as via a convolutional neural network (CNN) is exploited for its powerful ability to express the relationship between the original features and the transformed features \cite{ref6,ref7}. Since DNNs are differentiable models, the key aspect of this method is parameter adjustment by backpropagation \cite{ref8}. However, one of the disadvantages of DNNs is that there are too many hyperparameters, all of which need to be set manually for different datasets. More importantly, some schemes give better performance than DNNs for certain applications \cite{ref9,ref10}, such as XGBoost \cite{ref11,ref12} or random forest \cite{ref13}. For example, in the data science competitions run by Kaggle, models based on decision trees have achieved better results than DNNs.

To tackle these issues, Zhou and Feng \cite{ref14} proposed a novel deep learning model named gcForest, which was based on decision trees. gcForest utilizes two important characteristics of DNNs: representation learning and in-model feature transformation. First, gcForest applies the classical machine learning method of the decision tree, thus realizing the function of layer-by-layer processing. Next, it adopts multi-grained scanning, which is effective for processing instances with high numbers of dimensions and spatial or sequential relationships between features \cite{ref15}. Unlike DNN, gcForest is a non-differentiable model that does not require backpropagation training, and it also has fewer parameters. More importantly, gcForest is robust to hyperparameter settings, which means that users can obtain excellent performance on many datasets by using the default hyperparameters. A large number of numerical experimental results show that gcForest gives better performance than DNNs for the default hyperparameters \cite{ref14}, and gcForest has therefore been applied in many fields. For example, Ren et al. \cite{ref16} proposed a deep forest algorithm for multiple instance learning, which achieved superior performance. Yang et al. \cite{ref17} proposed a multi-label learning deep forest technique, which employed measure-aware feature reuse and layer growth to solve a multi-label learning problem. Utkin et al. \cite{ref18} proposed a novel model called adaptive weighted deep forest, in which each training instance was given a weight at each cascade level of the model. Recently, gcForest has been applied in the classification of schizophrenia data \cite{ref19}, disease classification \cite{ref20} and cancer detection \cite{ref21,ref22,ref23}. gcForest has not only achieved great success in the medical field, but has also been applied in other domains such as remote sensing \cite{ref24,ref25}, facial age estimation \cite{ref26} and EEG processing \cite{ref27}.

The experimental results show that a model with more grains and a larger number of forests can achieve better accuracy. However, the actual results of the application are not ideal due to the constraints of time costs. There are two main reasons for this phenomenon: firstly, gcForest passes all instances through all levels of the cascade; and secondly, multi-grained scanning typically converts one (original) instance into hundreds or even thousands of new instances, producing a high-dimensional input for the cascade structure. To address these issues, Pang et al. \cite{ref28} proposed a model called gcForestcs with a confidence screening mechanism, the principle of which is as follows. At each cascade level, each instance is screened using a threshold based on confidence. If the confidence of an instance is larger than the threshold, the final prediction of the instance can be obtained at the current level; otherwise, the results will continue to be input to the next level for training to achieve better performance. The framework of gcForestcs is shown in Figure 1.

\begin{figure}[h]
	\centering
	\includegraphics[width=1\linewidth]{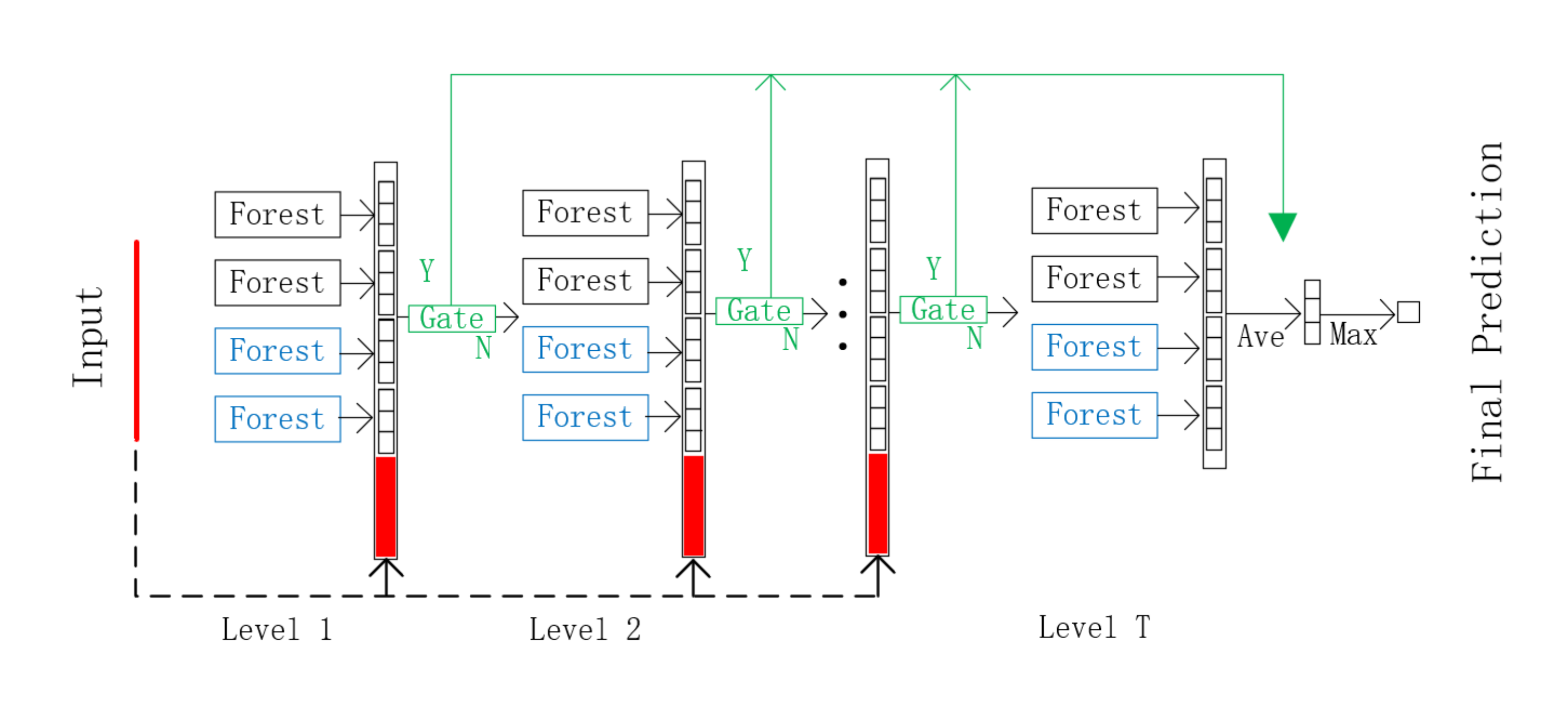}
	\caption{The framework of gcForestcs.}
	\label{fig:screenshot001}
\end{figure}

gcForestcs gives better performance in terms of training time than the original model on many datasets, and especially high-dimensional datasets such as those seen in image processing applications. However, gcForestcs relies more on the adjustment of hyperparameters than the original model, meaning that it has poorer prediction accuracy than the original model on many datasets when the default hyperparameters are used. The reason for this is that there is a group of instances with low accuracy in the high-confidence region, which refer to mis-partitioned instances. To address this issue, this paper proposes an algorithm called deep binning confidence screening forest, which adopts a strategy in which instances are binned based on their confidences. In this way, mis-partitioned instances can be detected. Our experiments show that for the same training hyperparameters, the DBC-Forest model gives better accuracy than gcForest and gcForestcs. More importantly, to achieve the same accuracy, DBC-Forest model is faster than the other models.

The rest of the paper is organized as follows. Section 2 briefly reviews the method used to produce the threshold in gcForestcs. Section 3 introduces the DBC-Forest model. Section 4 verifies the performance of this model on different datasets, and Section 5 summarises the contributions of this paper.

\section{Producing threshold of gcForestcs}

The confidence value, generated by the 3-fold cross-validation at each level, is used as a criterion to divide the data into two subsets in gcForestcs, and is the maximum value in the estimated class vector for the instance. For example, in a three-class classification problem, each level outputs a three-dimensional estimated class vector. If an estimated class vector is (0.6, 0.3, 0.1), then the confidence is 0.6. Each level of the cascade structure will generate a confidence value for each instance. If a certain level has $n$ instances, we obtain $(\bm{M_1},\bm{M_2},\dots \bm{M_n})$ confidences. The confidence $\bm{P_i}$ of the $i$-th $(1\le i\le n)$ instance $\bm{M_i}$ is:

\begin{equation}
		\bm{P_i}=max(\bm{M}_i \bm{x}_1,\bm{M}_i \bm{x}_2,\dots \bm{M}_i \bm{x}_c)
		\label{1}
\end{equation}
where $\bm{M_ix_c}$ is the $c$-th estimated class of $\bm{M_i}$. gcForestcs calculates the confidences of all the training instances, and sorts them in descending order based on their values.

\begin{equation}
	(\bm{P}_1,\bm{P}_2,\dots \bm{P}_n)=sort
	\left(\begin{array}{ccc}
		max(\bm{M}_1 \bm{x}_1,\bm{M}_1 \bm{x}_2,\dots \bm{M}_1 \bm{x}_c)\\
		\vdots\\
		max(\bm{M}_n \bm{x}_1,\bm{M}_n \bm{x}_2,\dots \bm{M}_n \bm{x}_c)\\
	\end{array}\right)
	\label{2}
\end{equation}

The threshold calculation in gcForestcs can be divided into two parts: ranking the instances and producing the threshold. The ranking of instances $(\bm{A}_1,\bm{A}_2,\dots \bm{A}_n )$ can be obtained by ranking the confidences  $(\bm{P}_1,\bm{P}_2,\dots \bm{P}_n)$. The production of threshold $Gate$ needs to calculate the accuracy of the regional instances. This region is obtained based on the confidence ranking of the instances. The upper bound of the region is the instance with the highest confidence, $\bm{A}_1$, and the lower bound slides from $\bm{A}_n$ to $\bm{A}_1$. gcForestcs calculates the accuracy of the region and compares it with the target accuracy $TA$.
\begin{equation}
      Gate=min\{\bm{P}_k|L_k\geqslant TA,k\in [1,n]\}
	\label{2}
\end{equation}
where $L_k=\frac{1}{k}\sum^k_{i=1}\mathbb{1}[p(\bm{A}_i)=y_i]$ is the accuracy of the instances $(\bm{A}_1,\bm{A}_2,\dots \bm{A}_k )$.\par
If the accuracy of $L_k$ is no less than $TA$, the confidence $\bm{P}_k$ of $\bm{A}_k$ is used as the threshold to divide all of the instances into two subsets: a high-confidence region  $Y$, and a low-confidence region $N$. The method used to calculate the threshold and process the division of instances is shown in Figure 2.
\begin{figure}[h]
	\centering
	\includegraphics[width=0.9\linewidth]{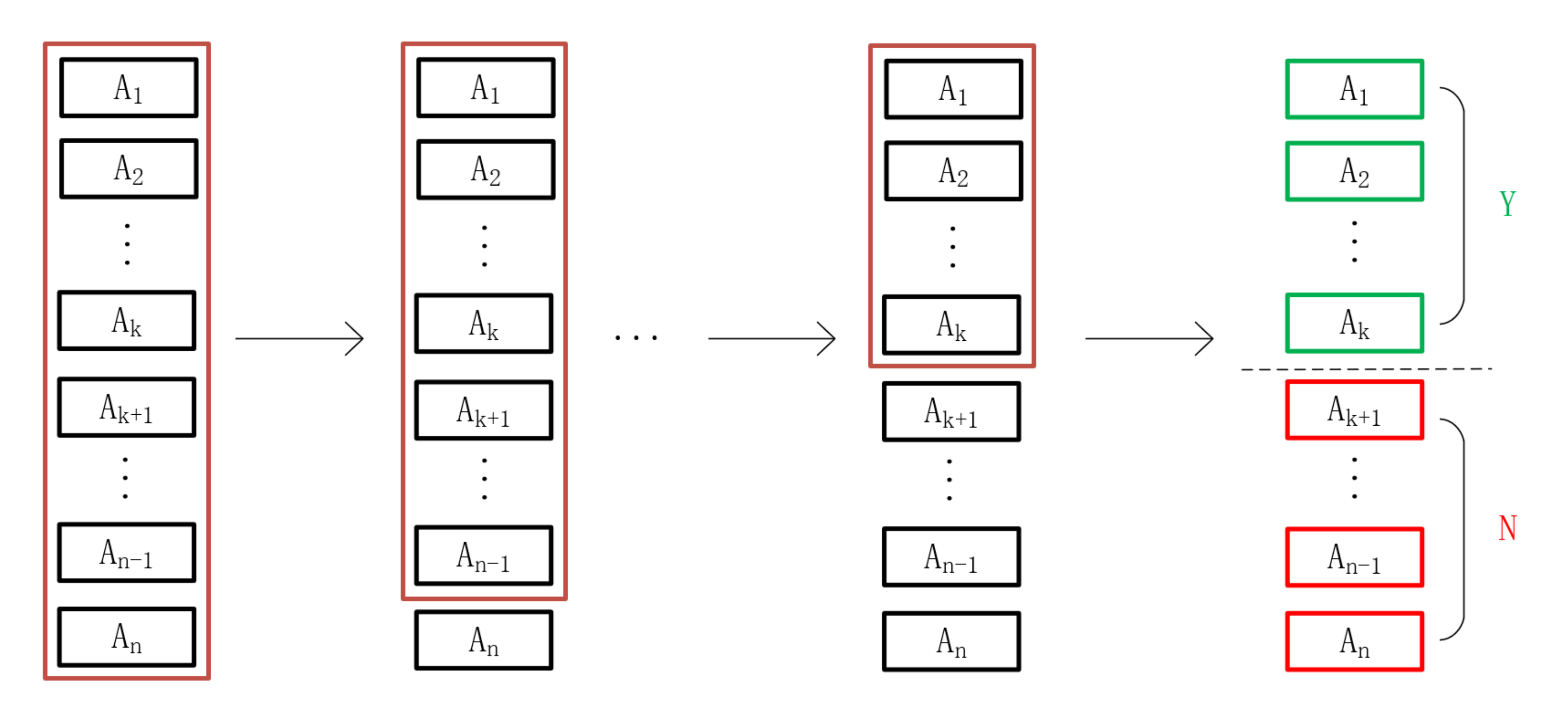}
	\caption{Threshold calculation and instance division processing: gcForestcs calculates the accuracy of the instances in the brown box region. $Y$ and $N$ are the high-confidence and low-confidence regions, respectively.}
	\label{fig:screenshot002}
\end{figure}

\section{DBC-Forest}

Section 3.1 describes the weakness of the method of producing the threshold in gcForestcs. The DBC-Forest approach used to find mis-partitioned instances is presented in Section 3.2, in which a binning strategy is employed to identify the threshold.



\subsection{Mis-partitioned Instances}

The method of producing the threshold in gcForestcs suffers from the issue of mis-partitioned instances. An illustrative example is given below. 

Example 1. Suppose there are 12 instances, as shown in Figure 3. The confidences are sorted in descending order. Of these, the predictions for instances 1, 2, 3, 4, 7 and 10 are correct, and the others are incorrect. Suppose the target accuracy $TA$ is 70\%. Since the accuracy of the high confidence region is 5/7 = 71.4\%, gcForestcs selects the confidence of instance 7 as the threshold. However, instances 5 and 6 are mis-partitioned, since they are incorrectly allocated to the high-confidence region. 

\begin{figure}[h]
	\centering
	\includegraphics[width=1\linewidth]{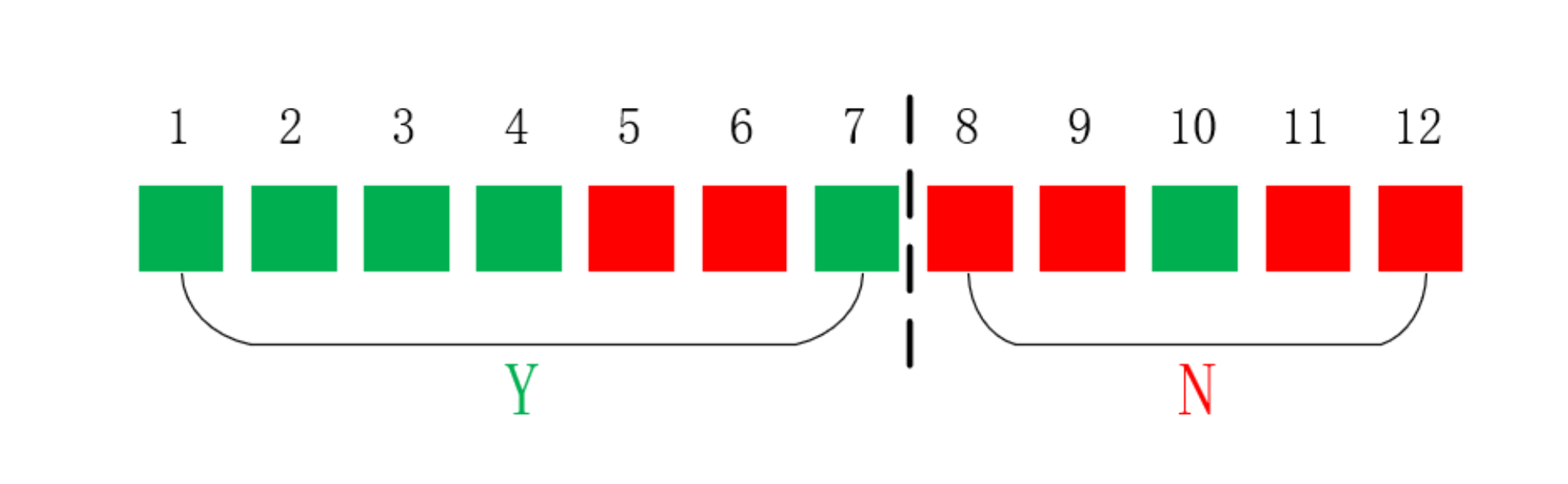}
	\caption{The result of threshold calculation according to gcForestcs. If $TA$=70\%, gcForestcs will select the confidence of instance 7 as the threshold, and instances 5 and 6 will be mis-partitioned.}
	\label{fig:screenshot003}
\end{figure}
Obviously, if the mis-partitioning of instances can be avoided, the prediction performance will be improved. Section 3.2 presents DBC-Forest, in which the threshold calculation method is improved.\par

\subsection{DBC-Forest}

The general framework of DBC-Forest is the same as that of gcForestcs, as shown in Figure 1. The difference between the two methods is that they adopt different strategies for producing the threshold. We now introduce the principle used to produce the threshold in DBC-Forest, which applies the following five steps to screen the instances at each level.

Step 1:  Rank the confidences to obtain $(\bm{A}_1,\bm{A}_2,\dots \bm{A}_n )$. This step is similar to the first step of gcForestcs.\par
Step 2: Set the bins $\bm{b}_1, \bm{b}_2 ,\dots  \bm{b}_k$ and place the instances ($\bm{A}_1,\bm{A}_2,\dots \bm{A}_n$ ) into the bins. Of these, instances $\bm{A}_1$, $\bm{A}_2$, \dots, and $\bm{A}_{n/k}$ with the highest confidence are put into $\bm{b}_1$, and the following $\ n/k$ instances are put into $\bm{b}_2$. Finally, the last $n/k$ instances with the lowest confidence are put into $\bm{b}_k$. Thus, each bin $\bm{b}_t  (1\leqslant t\leqslant k)$ can be obtained according to Equation (4).\par 
\begin{equation}
	(\bm{A}_{(\frac{n}{k})\times(t-1)+1},\bm{A}_{(\frac{n}{k})\times(t-1)+2},\dots \bm{A}_{(\frac{n}{k})\times t})\in \bm{b}_t 
\end{equation}

\begin{figure}[h]
	\centering
	\includegraphics[width=1\linewidth]{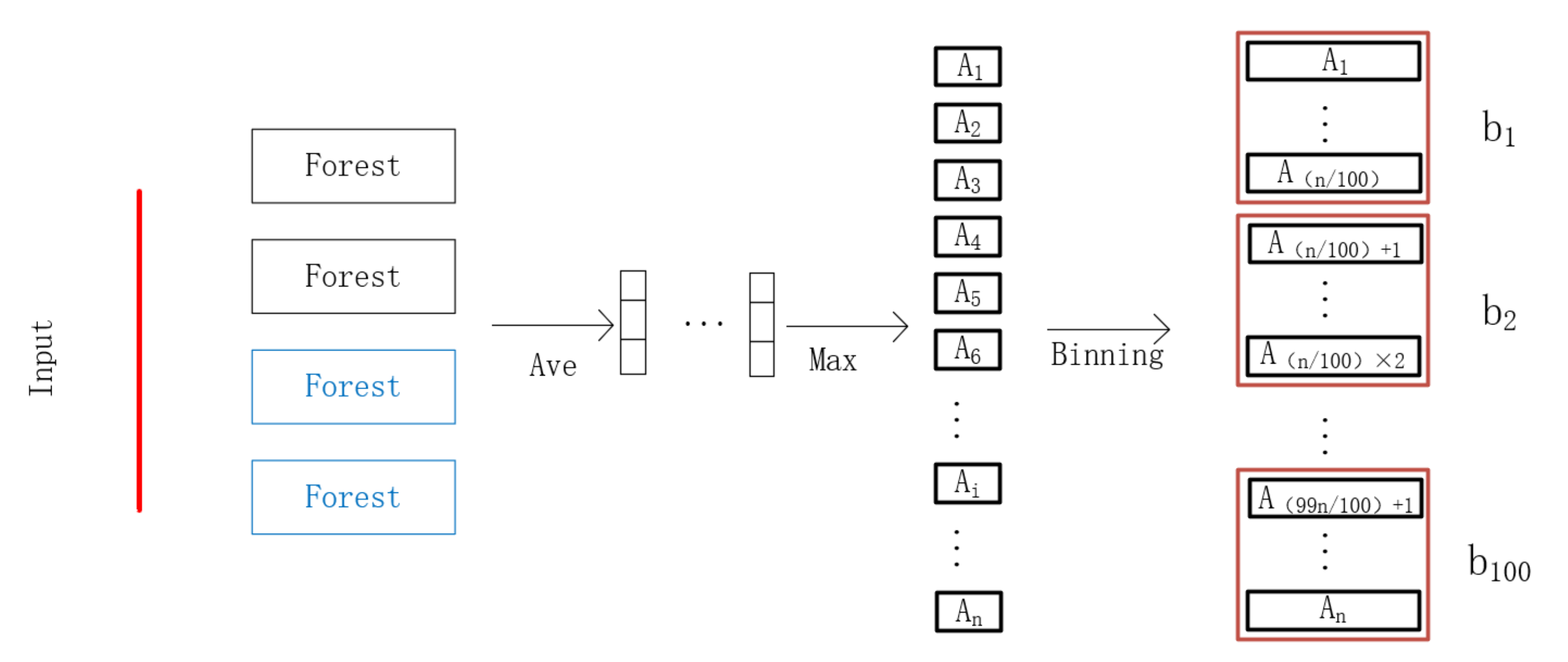}
	\caption{Binning process. For clarification, we set $k$=100.}
	\label{fig:screenshot004}
\end{figure}

For clarification, an example is shown in Figure \ref{fig:screenshot004}.

Step 3: Calculate the accuracy of the $n/k$ instances for each bin. The accuracy of bin $\bm{Pb}_t  (1\leqslant t\leqslant k)$ is the average accuracy of the instances in its bin, which can be calculated according to Equation (5):
\begin{equation}
		Pb_t = \frac{\sum_{i=\frac{n}{k}\times(t-1)+1}^{n\times \frac{t}{k}}\mathbb{1}[p(\bm{A_i})=y_i]}{n/k}
\end{equation}
where $p(\bm{A_i})$ and $y_i$ are the prediction and label of instance $\bm{A_i}$ , respectively. In this way, we can obtain the accuracies of all bins $(\bm{Pb}_1,\bm{Pb}_2,\dots \bm{Pb}_k )$, which are stored in $(\bm{b}_1,\bm{b}_2,\dots \bm{b}_k )$\par
Step 4: Produce the threshold $Gate$ according to $TA$. In DBC-Forest, the accuracy of each bin is compared with $TA$ from 1 to $k$, and the bin whose accuracy is less than $TA$ is found. If  $\bm{Pb}_{(j+1)} \le TA$, then the threshold $Gate$ is the confidence of instance $\bm{A}_{(b\times j/k)}$.

Step 5: Obtain the final result based on $Gate$. If the confidence of the instance is larger than  $Gate$, DBC-Forest will use the prediction of the current level as the final result $\bm{F_i}$, as shown in Equation (6); otherwise, the instance goes to the next level. \par
\begin{equation}
	\bm{F_i}=\{pre(\bm{A_i})|\bm{P_i}\geqslant Gate\}
\end{equation}\par
where pre($\bm{A}_i$) denotes the prediction of $\bm{A}_i$.\par

Example 2 illustrates the principle used to produce the threshold in DBC-Forest.  

Example 2. In this example, we use the same 12 instances as in Figure 3, and set $TA$=70\%. When the size of bins is two, there are six bins, as shown in Figure 5. We know that the accuracy of bin III is zero, which is less than $TA$. Hence, the confidence of instance 4 in bin II is selected as the threshold in DBC-Forest.

\begin{figure}[h]
	\centering
	\includegraphics[width=1\linewidth]{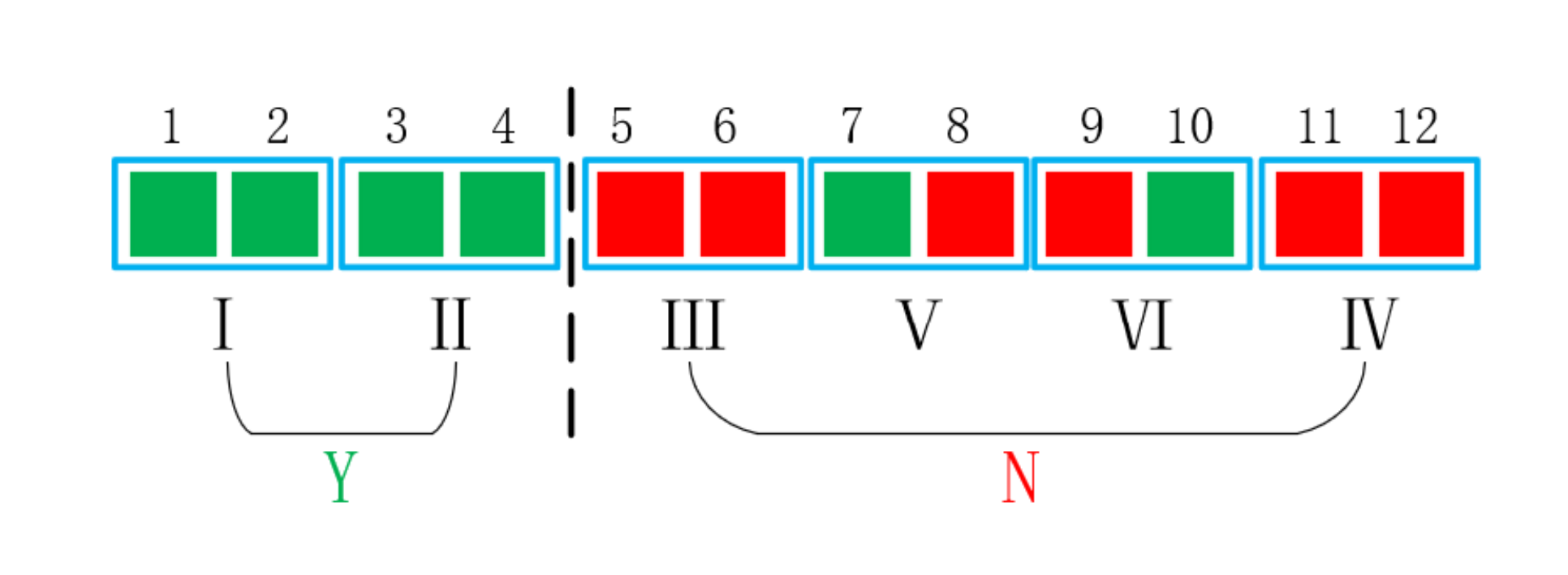}
	\caption{The result of threshold calculation in DBC-Forest. If $TA$=70\%, DBC-Forest selects the confidence of instance 4 as the threshold.}
	\label{fig:screenshot005}level
\end{figure}
Compared with gcForestcs, the advantage of DBC-Forest is that it can exactly generate the threshold and avoid mis-partitioned instances. But, it is worth noting that only in the case of unlimited depth of decision tree, the accuracies of all training instances are 100\%. In this case, the thresholds of both DBC-Forest and gcForestcs are 1, and the models of DBC-Forest and  gcForestcs are the same. This case can only happen when the dataset has few instances, since the confidence of each instance is generated by the 3-fold cross-validation at each level.The frame of DBC-Forest is shown in Figure 6.\par
\begin{figure*}[h]
	\centering
	\includegraphics[width=1\linewidth]{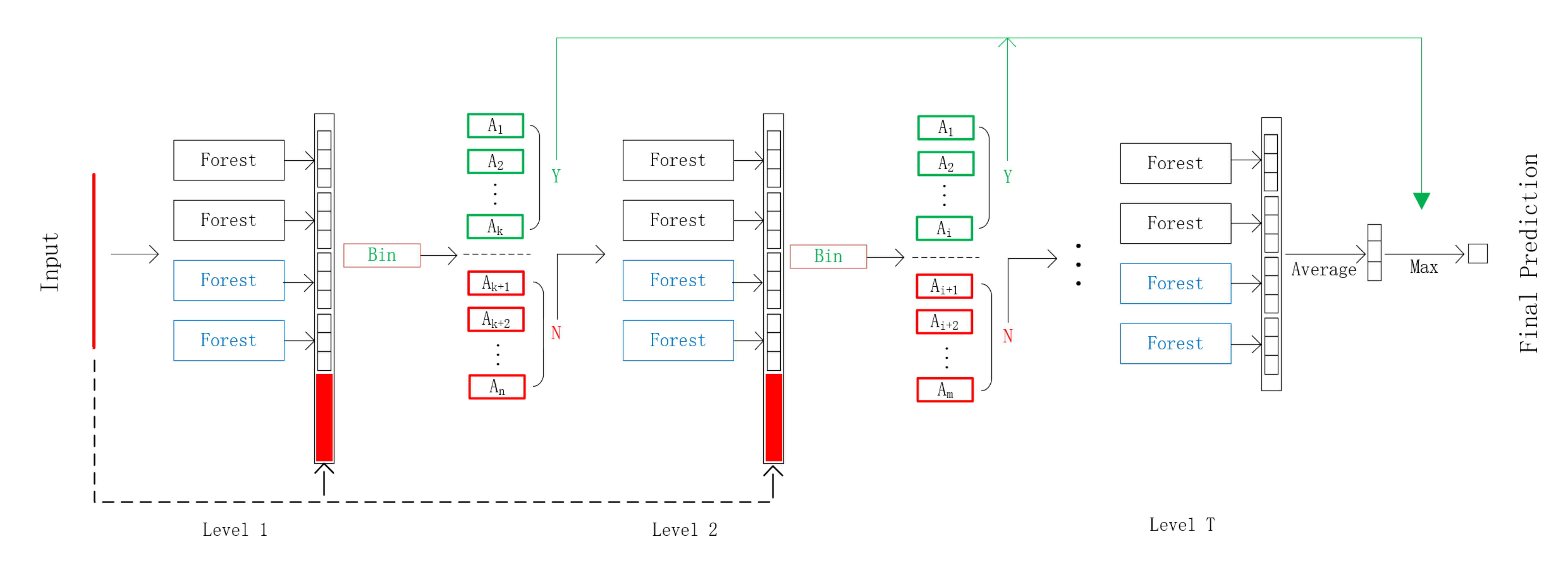}
	\caption{The framework of DBC-Forest}
	\label{fig:dbc-forest-frame}
\end{figure*}
\section{Experiments}

The goal of these experiments is to demonstrate that DBC-Forest can achieve better accuracy than other competitive models for the same hyperparameters, and that the training time is less than other models to achieve the same accuracy. Section 4.1 introduces the experimental environment and benchmark datasets. Section 4.2 explains the baseline methods and developed six research questions. Section 4.3 shows the influence of the size of bins. Section 4.4 reports the performance in terms of accuracy. Section 4.5 compares the thresholds of DBC-Forest and gcForestcs. Section 4.6 describes the performance of the algorithm in terms of robustness. Section 4.7 analyzes the efficiency of the binning confidence screening. Section 4.8 presents the performance in terms of training time.

\subsection{Experimental environment and benchmark datasets}
\textbf{Hardware:} In all experiments, we use a machine with 2×2.20 GHz CPUs (10 cores) and 128 GB main memory.

\textbf{Hyperparameter Settings:} To ensure a fair comparison, gcForest, gcForestcs and DBC-Forest use the same hyperparameters as follows. Cascade structure: Each level has a random forest and a completely random forest. The number of decision trees in the two forests is 50. The number of cascade levels stops increasing when the current level can not improve  the accuracy of the previous level for DBC-Forest, gcForestcs and gcForest, which can effectively reduce the time cost and complexity of the model. For gcForestcs and DBC-Forest, $TA$ is set to decrease the error rate by 1/2, and the size of the bin use in DBC-Forest is 100. For multi-grained scanning, the three window sizes are [d/4], [d/6] and [d/8], and the number of decision trees is 30. Multi-grained scanning can generate more features, which can highlight the efficiency of confidence and binning confidence scanning.

\textbf{Evaluation metrics:} In all experiments, we adopt the predictive accuracy as the classification performance measurement that is the most suitable for these balanced datasets. The training time is used to evaluate the efficiency. For fairness, we select the 5-fold cross-validation to verify the performance of DBC-Forest in all experiments.

\textbf{Datasets:} All of the datasets used in the experiments to verify the performance of DBC-Forest are representative datasets. MNIST, DIGITS, EMNIST \cite{ref29} and FASHION-MNIST \cite{ref30} are high-dimensional datasets that require multi-grained scanning. ADULT, BANK, YEAST, LETTER and IMDB are low-dimensional datasets which do not require multi-grained scanning.The IRIS is a dataset with few instances, which is used to verify the test performance of the proposed model when the confidence of each instance is 1.

\begin{table}[h]\footnotesize
	\caption{Description of datasets}
	\centering
	\begin{tabular}{ccccc}
		\hline
		Name                  & Train   & Test  &$D$   &$L$  \\ \hline
		MNIST                  & 56,000  & 14,000 & 784 & 10 \\
		DIGITS                & 1,437   & 360  & 64 & 10 \\
		EMNIST                 & 105,280 & 26,320 & 784 & 10 \\
		FASHION-MNIST       & 56,000  & 14,000 & 784 & 10 \\
		ADULT               & 39,074 & 9,768 & 14  & 2  \\
		BANK(BANK   MARKETING) & 32,950  & 8,238  & 20  & 2  \\
		YEAST                 & 1,187   & 297   & 8   & 10 \\
		LETTER             & 16,000  & 4,000  & 16  & 26 \\  
		IMDB               & 40,000  & 10,000  & 5,000  & 2 \\ 
		IRIS            & 120  & 30  & 4  & 3 \\\hline
	\end{tabular}
\end{table}

\subsection{Baseline methods and reaserch questions}

To validate the performance of DBC-Forest, six methods are used. The principles of them are described as follows.

\textbf{XGBoost:} XGBoost \cite{ref11}, as an ensemble learning algorithm improved by GBDT, introduces a regularized objective, and improves the loss function.

\textbf{LightGBM:} LightGBM \cite{ref31} is a novel GBDT algorithm, which contains two mechanisms: gradient-based one-side sampling and exclusive feature bundling to deal with large number of data instances and large number of features, respectively.

\textbf{mgrForest:} mgrForest \cite{ref32} is a multi-dimensional multi-grained residual forest algorithm, which maps feature vectors to higher levels for prediction.

\textbf{AWDF:}  AWDF \cite{ref18} is an improved deep forest method, which adopts the adaptive weight of every training instance at each cascade level.

\textbf{gcForest:} gcForest adopts multi-grained scanning and a cascade structure. Multi-grained uses a high-dimensional data cascade structure via layer-by-layer to output finally prediction.

\textbf{gcForestcs:} gcForestcs is an improved deep forest method. At each level, a threshold is automatically generated to screen the instances to reduce the time cost.

To validate the performance of DBC-Forest, we developed the following six research questions (RQ).

RQ 1: What the influence of size of bin has on the performance of the DBC-Forest?

RQ 2: Compared with state-of-the-art alternatives, what is the performance of DBC-Forest?

RQ 3: What is the difference between the thresholds in gcForestcs and DBC-Forest?

RQ 4: How does DBC-Forest performance under different hyperparameters?

RQ 5: Compared with gcForest, what is the effect of DBC-Forest at each level?

RQ 6: How is the running time of DBC-Forest?

To answer RQ 1, we adopted some different sizes of bin to validate the performance of DBC-Forest in Section 4.3. To address RQ2, we compared the performance of XGBoost, LightGBM, mgrForest, AWDF, gcForest, gcForestcs, and DBC-Forest  in Section 4.4. In response to RQ3, we explored the thresholds used in the binning confidence screening mechanism and confidence screening confidence mechanism to show different  in Section 4.5. For RQ 4, we set general hyperparameters to compare the performance of gcForest, gcForestcs, and DBC-Forest  in Section 4.6. To answer RQ 5, we compared the accuracy and the number of instances between gcForest and DBC-Forest at each level  in Section 4.7. To address RQ 6, we compared the time cost of gcForest, gcForestcs, and DBC-Forest under achieving similar accuracy  in Section 4.8.

\subsection{Influence of size of bins}

In this subsection, to report the influence of the size of bins on the model performance, we conduct experiments on FASHION-MNIST, BANK, LETTER and IMDB datasets. We use the size of bins varying from 10 to 140 to validate the performance of DBC-Forest. 
\begin{figure}[h]
	\centering 
	\subfigure[On FASHION-MNIST dataset]{ 
		\begin{minipage}{4.2cm}
			\includegraphics[width=4.2cm,height=3cm]{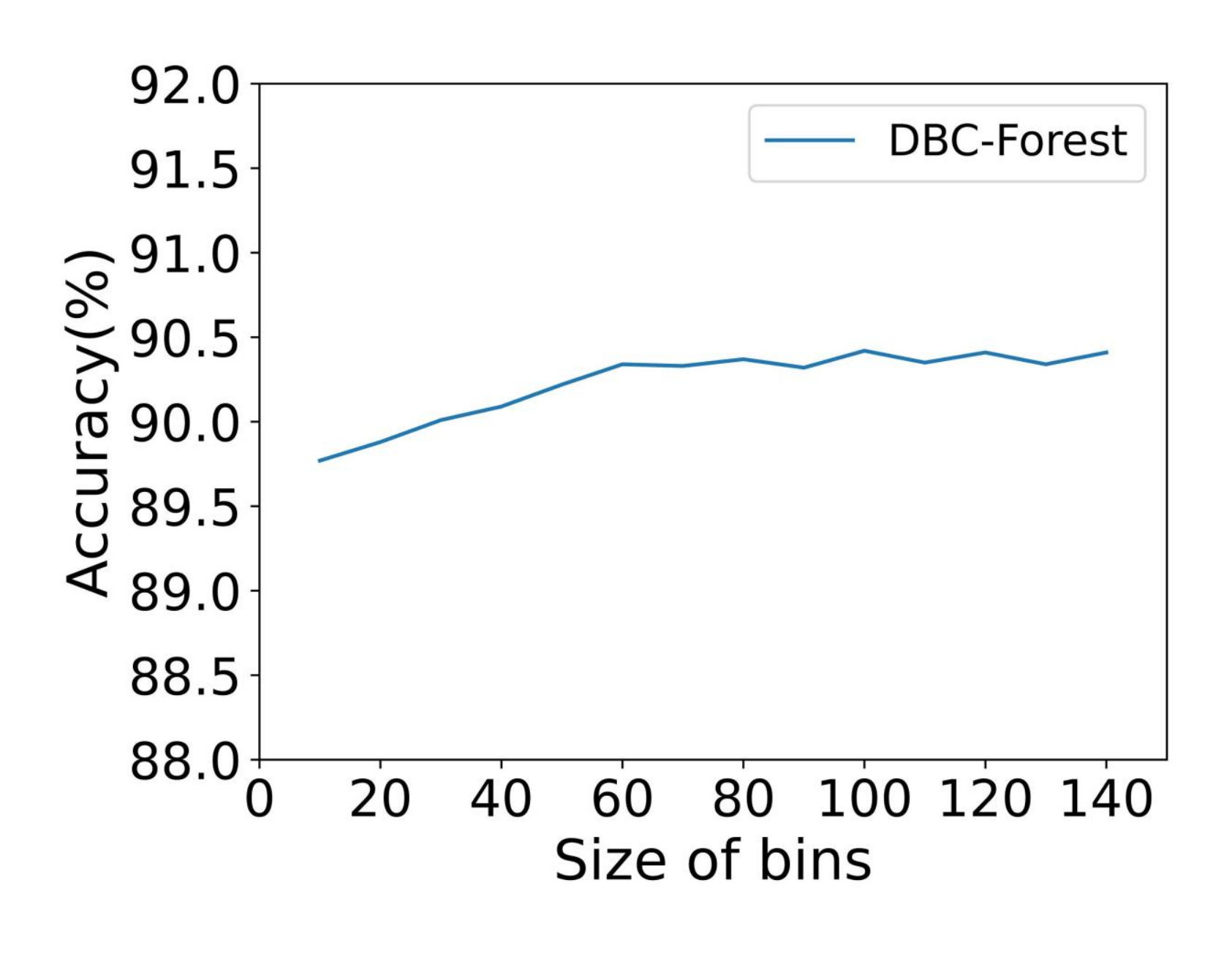}
		\end{minipage}
	}
	\subfigure[ On BANK dataset ]{ 
		\begin{minipage}{4.2cm}
			\includegraphics[width=4.2cm,height=3cm]{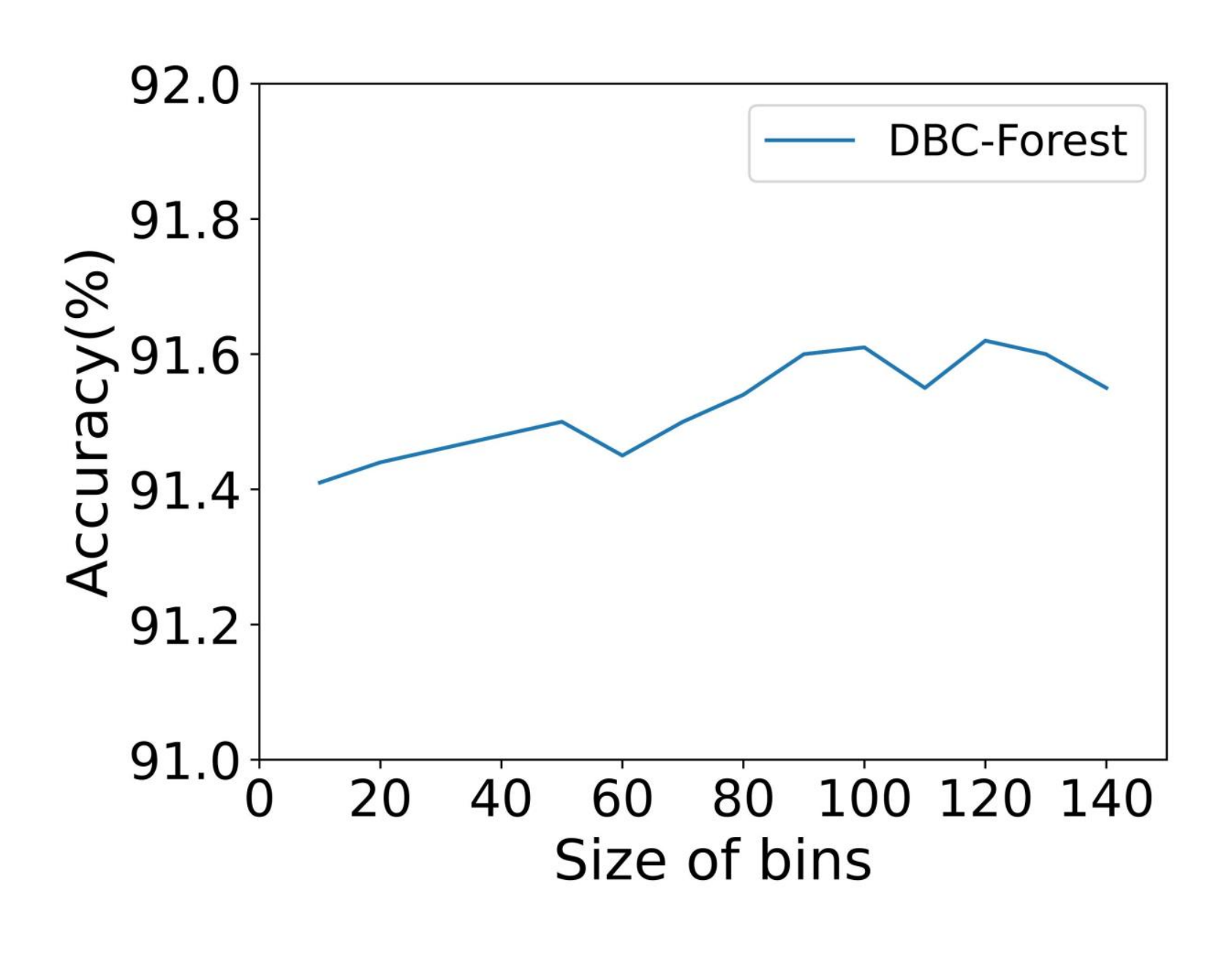} 
		\end{minipage}
	}
	\subfigure[On LETTER dataset]{ 
		\begin{minipage}{4.2cm}
			\includegraphics[width=4.2cm,height=3cm]{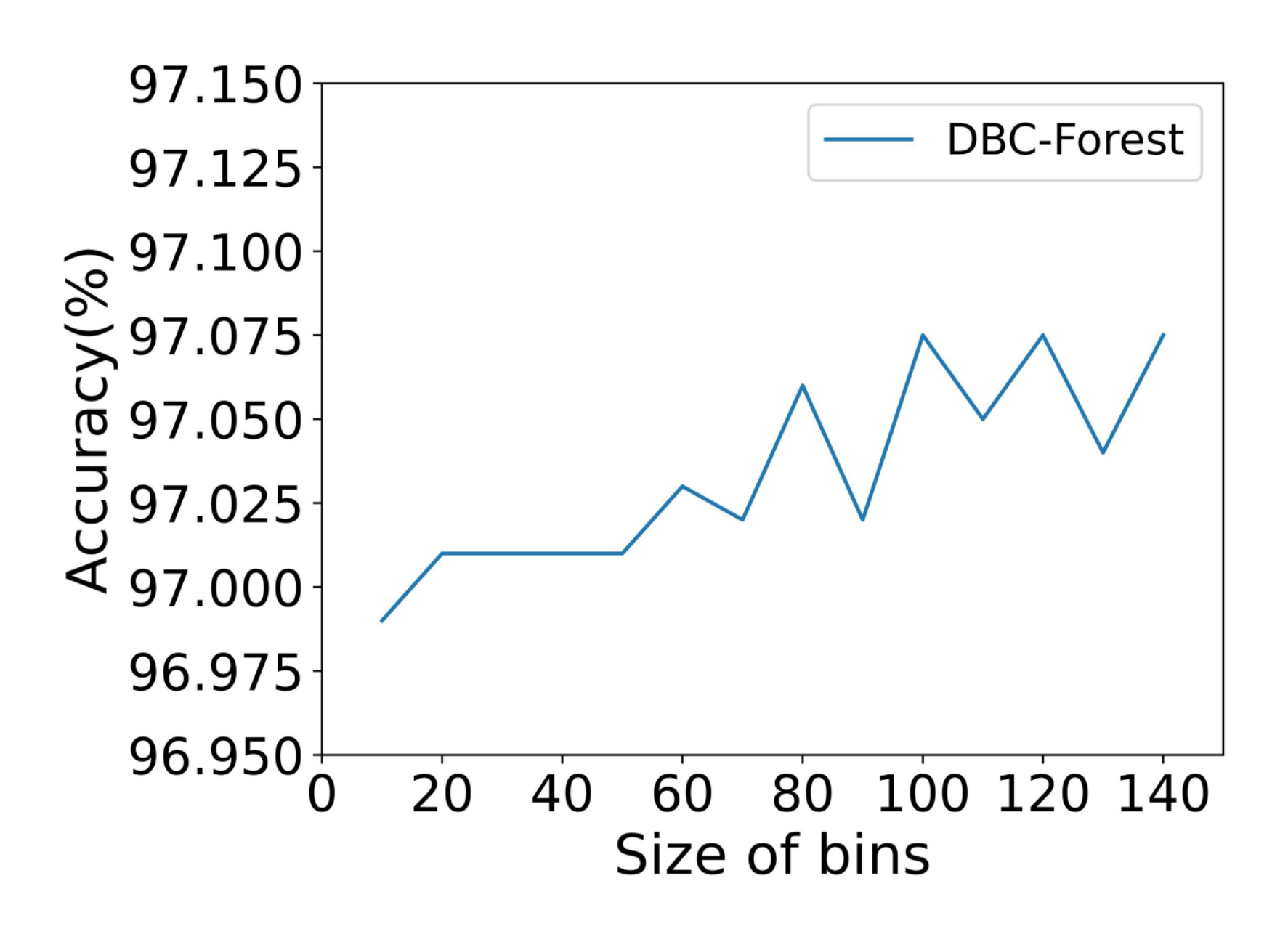} 
		\end{minipage}
	}
	\subfigure[On IMDB dataset]{ 
		\begin{minipage}{4.2cm}
			\includegraphics[width=4.2cm,height=3cm]{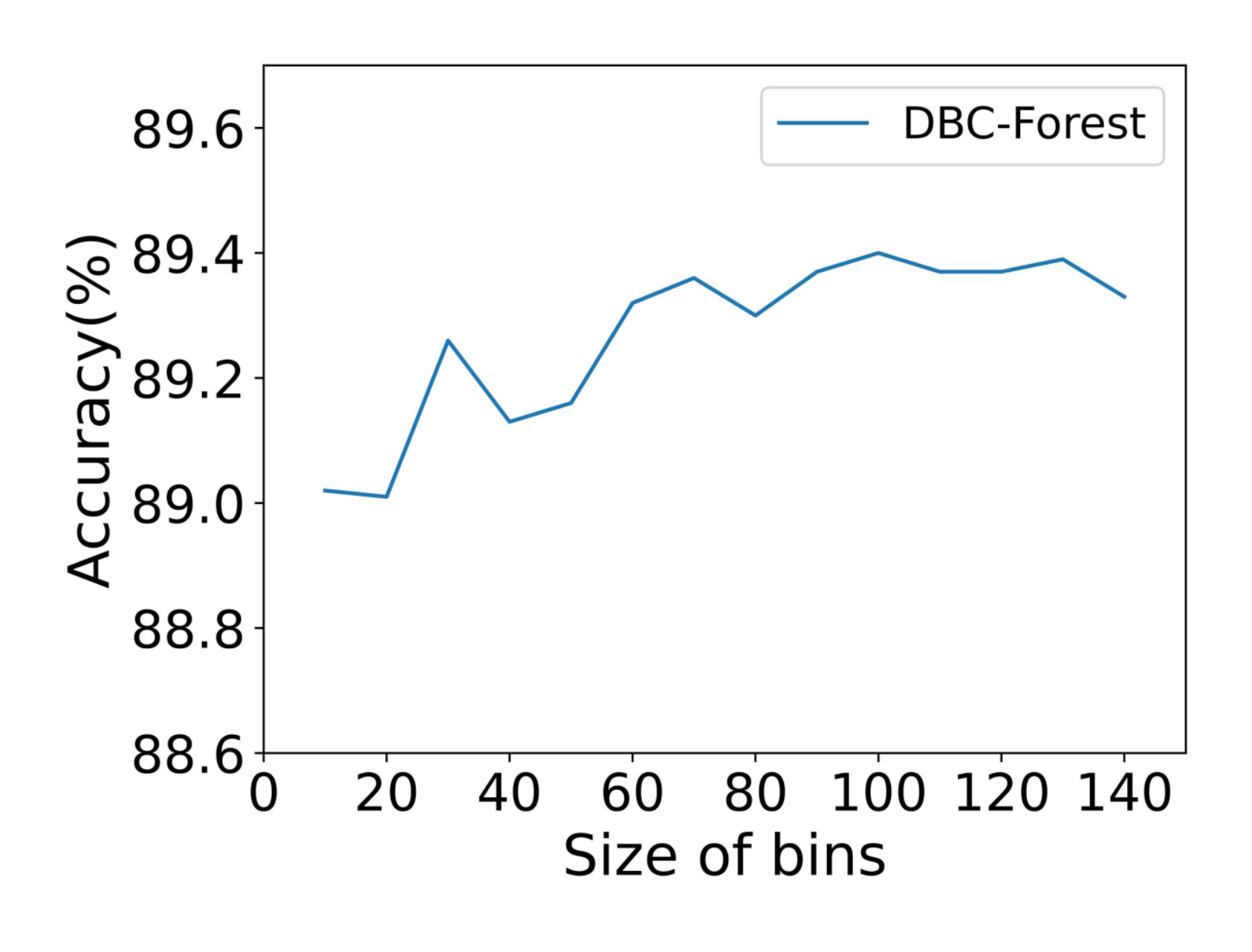} 
		\end{minipage}
	}
	\caption{Influence of size of bins} 
\end{figure}

Experiments show that the model performance varies with the size of bins, but the fluctuations are small. For example, as Figure 7 (a) shows, the difference between the lowest accuracy and the highest accuracy is only 0.64\%. More importantly, when the size of bins is 100, the accuracy reaches the best accuracy 90.42\%, and the same phenomenon can be found on other datasets. Hence, the size of bins is selected 100 in the rest experiments.
\subsection{{Accuracy}}

In this section, we compare the performance of DBC-Forest with other classical methods in terms of accuracy. The hyperparameters of XGBoost are the max depth and learning rate, which are 3 and 0.1, respectively. The hyperparameters of LightGBM are the learning rate and number of leaves which are 0.1 and 31, respectively. The default parameters of AWDF, mgrForest, gcForest, gcForestcs and DBC-Forest are given in Subsection 4.1. The experimental results are shown in Table 2.


\begin{table*}[!h]
	\caption{Comparison of accuracy (\%)}
	\centering
	\setlength{\tabcolsep}{1.3mm}{
		\begin{tabular}{cccccccc}
			\hline
			& XGBoost & LightGBM & mgrForest & AWDF  & gcForest & gcForestcs & DBC-Forest \\ \hline
			\multicolumn{1}{c}{MNIST}  & 97.75$\pm$0.17   & 97.58$\pm$0.17    & 97.61$\pm$0.19    & 98.89$\pm$0.11 & 98.77$\pm$0.12   & 98.36$\pm$0.14 &\textbf{99.03$\pm$0.07} \\
			\multicolumn{1}{c}{DIGITS} & 96.83$\pm$0.67   & 97.66$\pm$0.28  & 95.54$\pm$0.75  & \textbf{98.23$\pm$0.60} & 97.72$\pm$0.71 & 97.38$\pm$0.57      & 97.88$\pm$0.67\\
			\multicolumn{1}{c}{EMNIST} & 81.05$\pm$0.20   & 66.45$\pm$11.33    & 81.41$\pm$0.18   & 86.74$\pm$0.18 & 86.55$\pm$0.20 & 87.24$\pm$0.23    & \textbf{87.32$\pm$0.29} \\
			\multicolumn{1}{c}{FASHION-MNIST}& 90.30$\pm$0.33 & 90.07$\pm$0.25 & 87.71$\pm$0.34  & 89.89$\pm$0.27 & 89.99$\pm$0.33  & 89.94$\pm$0.29 & \textbf{90.57$\pm$0.22} \\
			\multicolumn{1}{c}{ADULT} & 87.05$\pm$0.11   & \textbf{87.45$\pm$0.24}    & 85.35$\pm$0.36     & 85.86$\pm$0.15 & 85.99$\pm$0.37 & 86.04$\pm$0.35 & 86.11$\pm$0.29 \\
			\multicolumn{1}{c}{BANK} & 91.48$\pm$0.20   & \textbf{91.77$\pm$0.26}    & 91.41$\pm$0.71    & 91.45$\pm$0.23 & 91.43$\pm$0.23 & 91.53$\pm$0.19 & 91.62$\pm$0.16  \\
			\multicolumn{1}{c}{YEAST}  & 59.63$\pm$2.22   & 57.40$\pm$3.86     & 58.55$\pm$3.14     & \textbf{62.53$\pm$4.10} & 62.06$\pm$3.77    & 62.00$\pm$3.5  & 62.13$\pm$3.70 \\
			\multicolumn{1}{c}{LETTER}  & 96.30$\pm$0.42   & 96.73$\pm$0.39 & 92.18$\pm$0.38 & 96.65$\pm$0.18 & 97.02$\pm$0.22  & 96.83$\pm$0.27  & \textbf{97.07$\pm$0.23} \\
			\multicolumn{1}{c}{IMDB} & 86.43$\pm$0.14   & 86.56$\pm$0.29    & 83.20$\pm$0.54  & 88.94$\pm$0.23 & 88.81$\pm$0.10 & 88.88$\pm$0.14  & \textbf{89.39$\pm$0.32} \\ \hline
	\end{tabular}}
\end{table*}

Table 2 shows that the gcForest-based model achieves the best performance on most datasets. For example, DBC-Forest achieves the highest accuracy of 90.57\% on FASHION-MNIST dataset. This shows that multi-grained scanning has advantages in the processing of high-dimensional datasets. More importantly, the  performance of DBC-Forest indicates that the binning confidence screening method improves the accuracy of the model.

To further evaluate the performance of DBC-Forest, we employ $t$-Test, Friedman Test, and Nemenyi Test for statistical hypothesis test.

1. $t$-Test: The most commonly utilized in the machine learning $t$-Test are the paired $t$-Test and the 5$\times$2 fold cross-validation paired $t$-test \cite{ncom1998}. Although some studies show that the 5$\times$2 fold cross-validation paired $t$-Test is better than the paired $t$-Test, we select paired $t$-Test in this paper. The reason is as follows. First, the 5$\times$2 fold cross-validation paired $t$-Test needs five times 2-fold cross-validation. However, we adopt one times 5-fold cross-validation in experiments. Second, although the paired $t$-Test has flaws, Steven \cite{TKDD1997} shows that it is difficulty agreeing on the correct framework for hypothesis testing in complex experimental designs . For example, the 5$\times$2 fold cross-validation paired $t$-Test has a much higher Type II error than paired $t$-Test \cite{TKDD2004}. Third,  we also adopt Friedman Test and Nemenyi Test for further statistical hypothesis test. Hence, in this paper, we adopt the paired t-Test to evaluate the performance DBC-Forest.

The processing of paired $t$-Test is shown as follows. First, we compare two model accuracies and get their different value $D$. We then obtain the statistic according to $\left| \mu/\sqrt{\frac{1}{kr}\sigma^{2}} \right|$ \cite{TKDD2004}, where $k$ is the fold of cross-validation times, $r$ is the $k$-fold cross-validation times, $\mu$ is the mean of $D$, $\sigma$ is the standard deviation of $D$. Since we adopt one times 5-fold cross-validation, the paired $t$-Test statistic is $\left| \sqrt{5}\mu/\sigma \right|$. Final, we compare $T_{0.05,4}$ with the statistics to accept (N) or reject (Y) the null hypothesis. The results of paired $t$-Test results are shown in Table 3.

\begin{table*}[!h]
	\caption{Significance test of $t$-Test}
	\centering
	\setlength{\tabcolsep}{1.3mm}{
		\begin{tabular}{ccccccc}
			\hline
			& XGBoost & LightGBM & mgrForest & AWDF  & gcForest & gcForestcs \\ \hline
			\multicolumn{1}{c}{MNIST}  & Y (21.770)  & Y (20.893)    & Y (18.674)     & Y (2.158) & Y (6.340)   & Y (10.328)  \\
			\multicolumn{1}{c}{DIGITS} & Y (2.906)   & N (1.239)  & Y (6.454)  & Y (3.110) & Y (0.986)  & Y (3.177)    \\
			\multicolumn{1}{c}{EMNIST} & Y (64.190)  & Y (4.022)    & Y (67.231)   & Y (14.143) & N (1.268) & Y (18.438)    \\
			\multicolumn{1}{c}{FASHION-MNIST}& Y (2.196) & Y (6.436)& Y (30.176)  & Y (8.597) & Y (7.323)  & Y (14.350)  \\
			\multicolumn{1}{c}{ADULT} &Y (4.945)   & Y (10.965)    & Y (3.409)     & Y (2.621) & N (1.372) & N (1.289)  \\
			\multicolumn{1}{c}{BANK} & N (0.878)   & N (1.164)    & Y (2.461)    & N (1.100) & Y (2.689) & Y (3.477)   \\
			\multicolumn{1}{c}{YEAST}  & Y (2.368)   & N (1.779)     & Y (2.484)     & N (0.183) & N (1.118)    & N (0.246)  \\
			\multicolumn{1}{c}{LETTER}  & Y (6.956)   & Y (3.805) & Y (72.742)  & Y (15.096) & N (0.964)  & Y (5.827)   \\
			\multicolumn{1}{c}{IMDB} & Y (23.278)   & Y (35.757)   & Y (33.717)  & Y (4.902) & Y (4.042) & Y (5.457)   \\ \hline
	\end{tabular}}
\end{table*}

As shown in Table 3, DBC-Forest is significantly different from other models on most datasets. For example, on LETTER dataset, the 5-fold cross-validation accuracies of DBC-Forest are 97.33, 96.97, 96.72, 97.33, and 97.00, and the 5-fold cross-validation accuracies of gcForestcs are 97.22, 96.60, 96.47, 97.03, and 96.83. Then the differnt value $D$ = (97.33-97.22, 96.97-96.60, 96.72-96.47, 97.33-97.03, 97.00-96.83) = (0.11, 0.37, 0.25, 0.30, 0.17). Thus, $\mu$ = 0.092, and $\sigma$ = 0.24, and the statistic is  $\left| \sqrt{5}\mu/\sigma\right|$ = $\left| \sqrt{5}\times0.24/0.092 \right|$ = 5.827. Finally, we reject the null-hypothesis that DBC-forest is equal to gcForestcs, since the statistic is larger than $T_{0.05,4}$ =2.132. However, we accept the null-hypothesis that DBC-forest is equal to gcForest on LETTER dataset, the processing is as follows. The 5-fold cross-validation accuracies of gcForest are 97.12, 97.00, 96.62, 97.28, and 97.10, and $D$ = (0.21, -0.03, 0.10, 0.05, -0.10).  Finally, the paired $t$-Test statistic is $\left| \sqrt{5}\mu/\sigma\right|$ = $\left| \sqrt{5}\times0.046/0.107 \right|$ = 0.964, and larger than $T_{0.05,4}$ = 2.132.

2. Friedman Test and Nemenyi Test: To compare the performance of all models, we adopt the Friedman Test \cite{TYCB2021} for significance test. For each dataset, we rank all models according to their accuracy. The highest is 1, and the second highest is 2, so on and so forth. Therefore, the smaller the mean value is, the better the performance is. The mean value $r$ of XGBoost, LightGBM, mgrForest, AWDF, gcForest, gcForestcs, and DBC-Forest are 4.67, 4.33, 6.55, 3.44, 3.78, 3.67, 3.67, and 1.56, respectively. Then we conduct the Friedman Test according to $(N-1)\tau_{\chi^2}/(N(k-1)\tau_{\chi^2})$, where $\tau_{\chi^2}=12N(\sum_{i=1}^{k}r_i^2-k(k+1)^2/4)/k(k+1)$, $N$ and $k$ are the number of datasets and the number of models, respectively. In the end, we reject the null-hypothesis, since the statistic is 7.481 which is greater than $F_{0.05,(6,48)}$ = 2.295.

To further differentiate the algorithms, we utilize the Nemenyi Test. The critical difference value of the Nemenyi Test is 2.742 according to  $q_{0.1,7}\sqrt{k(k+1)/6N}$, where  $q_{0.1,7}$ = 2.693. As shown in Figure 8, DBC-Forest is significantly better than most models.
\begin{figure}[h]
	\centering
	\includegraphics[scale=0.42]{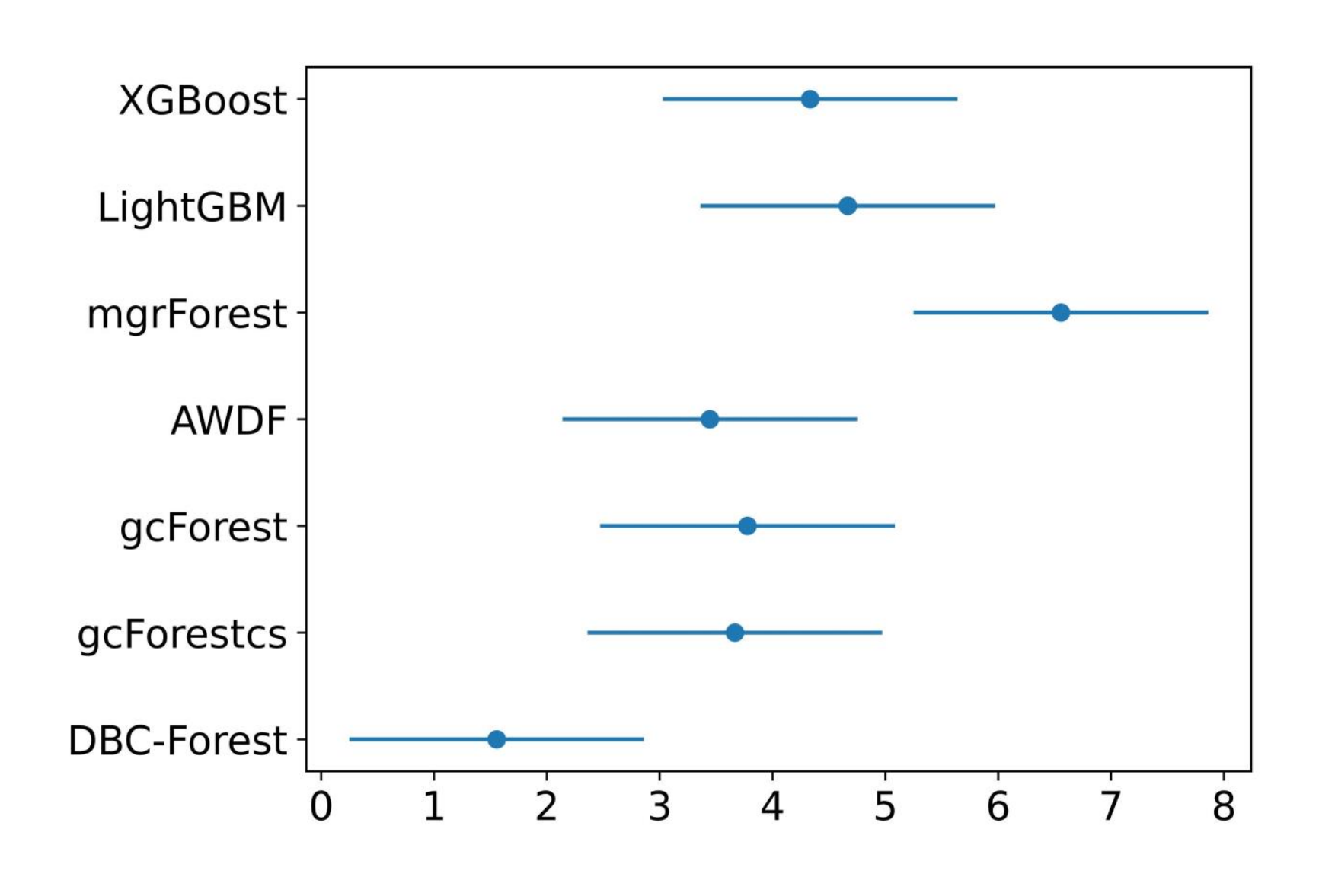}
	\caption{Nemenyi Test figure of model accuracy. The dots are the mean rank of models, and the horizontal bar across each dot donates critical difference.}
\end{figure}

\subsection{Comparison of thresholds}

DBC-Forest is an improved model of gcForestcs. Thus, gcForestcs is selected in this subsection. To show the difference of the thresholds of the two models, we select MNIST and BANK datasets with many instances, and IRIS dataset with few instances. Since the two models generate a large number of levels, we only show the comparison of the first level. The results are shown in Figures 9, 10 and 11, respectively.
\begin{figure}[h]
	\centering
	\includegraphics[scale=0.42]{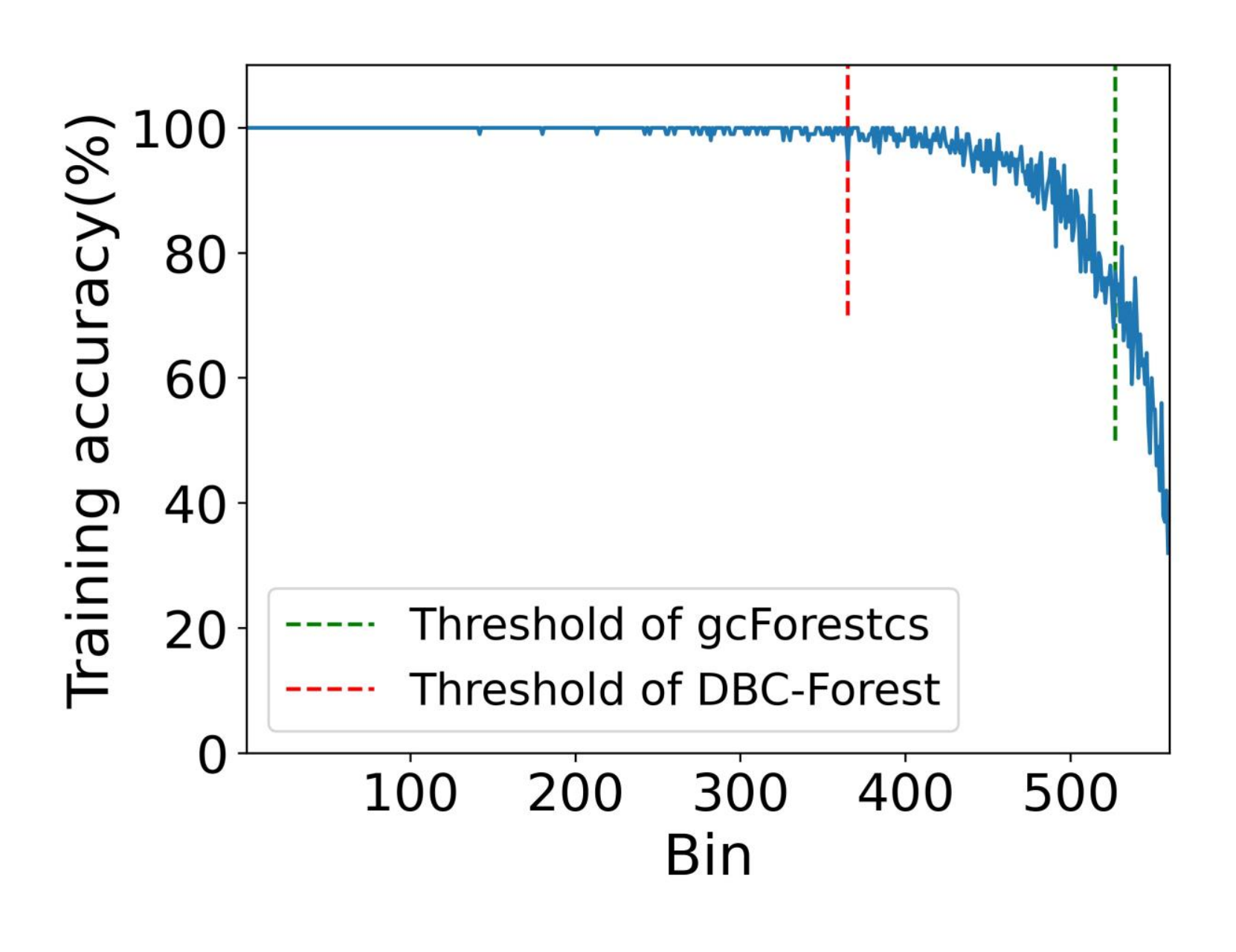}
	\caption{Comparison of thresholds on MNIST dataset}
\end{figure}
\begin{figure}[h]
	\centering
	\includegraphics[scale=0.42]{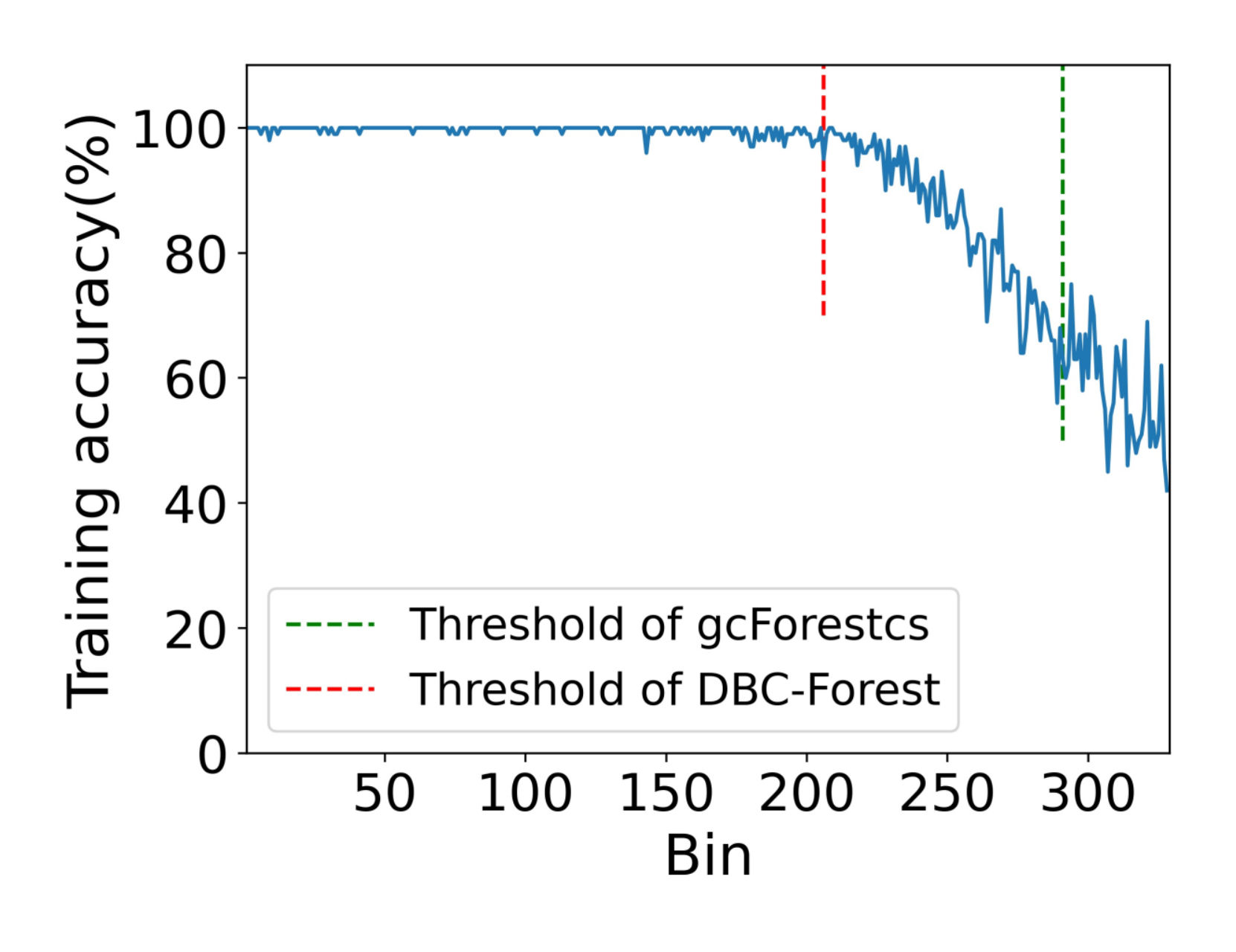}
	\caption{Comparison of thresholds on BANK dataset}
\end{figure}
\begin{figure}[h]
	\centering
	\includegraphics[scale=0.42]{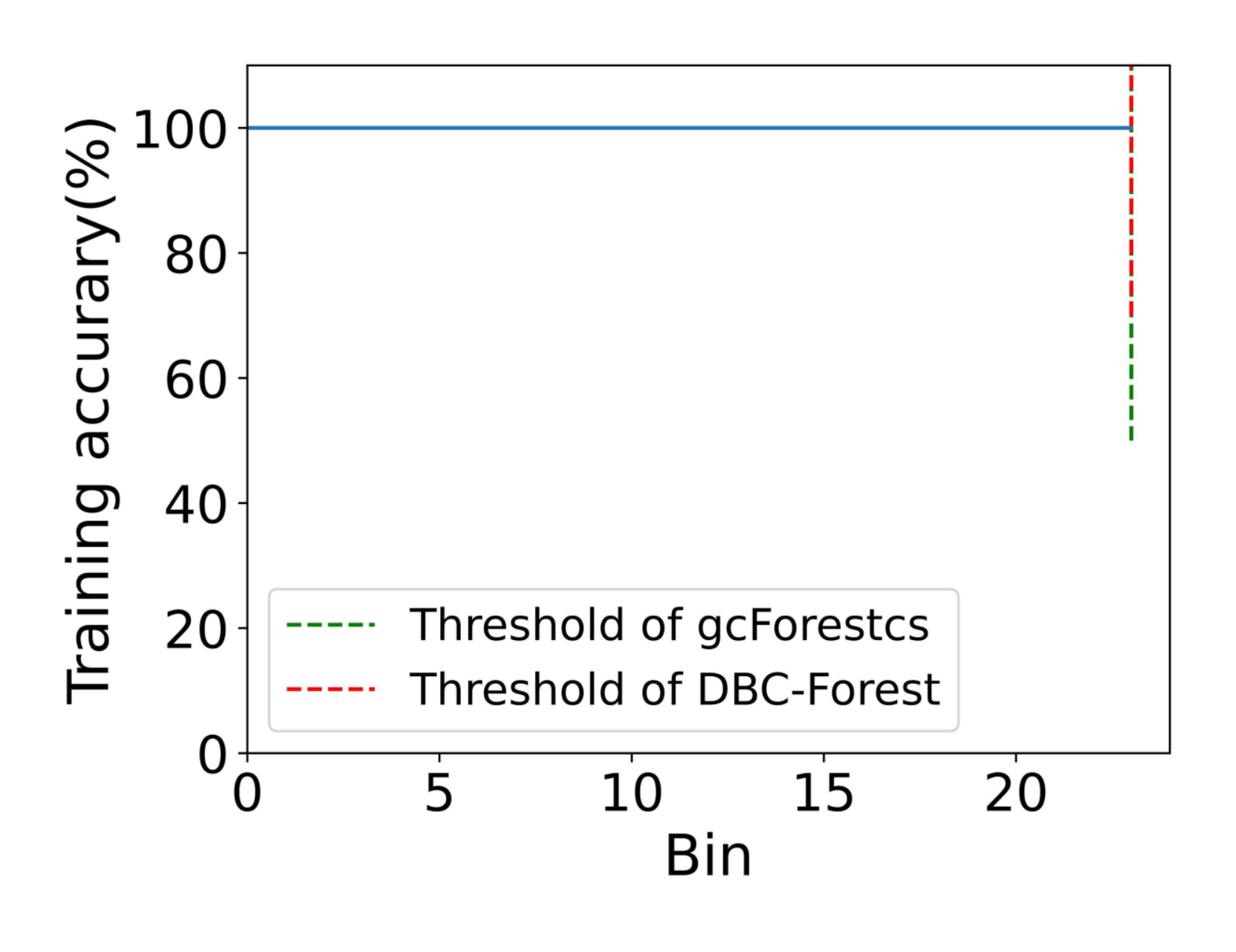}
	\caption{Comparison of thresholds on iris dataset}
\end{figure}

Figures 9 and 10 show that the thresholds of DBC-Forest are more reasonable than those of gcForestcs, since gcFoerstcs screens some instances with low accuracy. For example, on MNIST dataset, gcFoerstcs screens the instances in the first 527 bins, while DBC-Forest does in the first 365 bins. The instances from the 365th to 526th bins are mis-partitioned, which are incorrectly allocated to the high-confidence region.

Figure 11 shows that the two thresholds are the same. In this case, the training errors of the two models are both 0, i.e., there is no mis-partitioned instances. However, it should be noted that the test accuracies of the two models on IRIS dataset are both 93.33\%, which means that the generalization abilities of the two models are weak.

\subsection{{Robustness}}

Our goal is to validate the accuracy of DBC-Forest and its robustness (here, robustness means that the model always achieves excellent performance for different parameters). Three models are compared: gcForest, gcForestcs and DBC-Forest. The experiment is divided into two parts, with and without multi-grained scanning. The robustness of a model can be evaluated based on whether the model depends on the adjustment of the parameters. If a model works well with suitable parameters but the accuracy decreases rapidly with more general parameters, then the model depends on the adjustment of the parameters. To determine the robustness of the model, the experiment with multi-grained scanning uses two different hyperparameters: default parameters and general parameters. The general hyperparameters for multi-grained scanning are as follows: the three window sizes were [d/2], [d/3], [d/4], and the number of decision trees in the forest is 10. Comparisons of the default and general parameters, with and without multi-grained scanning, are shown in Tables 4, 5 and 6.

\begin{table}[h]
	Comparison of 5-fold cross-validation test accuracy with default parameters for high-dimensional datasets (\%)
	\centering
	\begin{tabular}{cccccc}
		\cline{1-4}
		& gcForest & gcForestcs & DBC-Forest &  &  \\ \cline{1-4}
		\multicolumn{1}{c}{MNIST} & 98.77$\pm$0.12 & 98.36$\pm$0.14 & \textbf{99.03$\pm$0.07} &  &  \\
		\multicolumn{1}{c}{DIGITS} & 97.72$\pm$0.71 & 97.38$\pm$0.57 & \textbf{97.88$\pm$0.67} &  &  \\
		\multicolumn{1}{c}{EMNIST} & 86.55$\pm$0.20 & 87.24$\pm$0.23 & \textbf{87.32$\pm$0.29} &  &  \\
		\multicolumn{1}{c}{FASHION-MNIST} & 89.99$\pm$0.33 & 89.94$\pm$0.29 & \textbf{90.57$\pm$0.22} &  &  \\ \cline{1-4}
	\end{tabular}
\end{table}
\begin{table}[h]
	\caption{Comparison of 5-fold cross-validation test accuracy with general parameters for high-dimensional datasets (\%)}
	\centering
	\begin{tabular}{cccccc}
		\cline{1-4}
		& gcForest & gcForestcs & DBC-Forest &  &  \\ \cline{1-4}
		\multicolumn{1}{c}{MNIST} & 98.19$\pm$0.13 & 97.12$\pm$0.19 & \textbf{98.63$\pm$0.08} &  &  \\
		\multicolumn{1}{c}{DIGITS} & 97.10$\pm$1.24 & 92.65$\pm$1.59 & \textbf{97.32$\pm$0.84} &  &  \\
		\multicolumn{1}{c}{EMNIST} & 84.39$\pm$0.24 & \textbf{86.39$\pm$0.25} & 85.31$\pm$0.29 &  &  \\
		\multicolumn{1}{c}{FASHION-MNIST} & 89.72$\pm$0.34 & 89.16$\pm$0.30 & \textbf{90.42$\pm$0.22} &  &  \\ \cline{1-4}
	\end{tabular}
\end{table}
\begin{table}[h]
	\caption{Comparison of 5-fold cross-validation test accuracy for low-dimensional datasets (\%)}
	\centering
	\begin{tabular}{cccccc}
		\cline{1-4}
		& gcForest & gcForestcs & DBC-Forest &  &  \\ \cline{1-4}
		\multicolumn{1}{c}{ADULT} & 85.99$\pm$0.37 & 86.04$\pm$0.35 & \textbf{86.11$\pm$0.29} &  &  \\
		\multicolumn{1}{c}{BANK} & 91.43$\pm$0.23 & 91.53$\pm$0.19 & \textbf{91.62$\pm$0.16} &  &  \\
		\multicolumn{1}{c}{YEAST} & 62.06$\pm$3.77 & 62.00$\pm$3.50 & \textbf{62.13$\pm$3.70} &  &  \\
		\multicolumn{1}{c}{LETTER} & 97.02$\pm$0.22 & 96.83$\pm$0.27 & \textbf{97.07$\pm$0.23} &  &  \\
		\multicolumn{1}{c}{IMDB} & 88.81$\pm$0.10 & 88.88$\pm$0.14 & \textbf{89.39$\pm$0.32} &  &  \\ \cline{1-5}
	\end{tabular}
\end{table}

The experimental results give rise to the following observations.

1. As shown in Table 4, the DBC-Forest model gives a higher accuracy than both gcForest and gcForestcs. For example, on FASHION-MNIST dataset, the accuracies of gcForest, gcForestcs and DBC-Forest are 90.57\%, 89.94\% and 89.99\%, respectively. This is due to the fact that the binning confidence screening improves accuracy. 

2. From Tables 4 and 5, it can be seen that the DBC-Forest model has better robustness performance than both the gcForest and gcForestcs models. For example, on MNIST dataset, it can be seen that the accuracies  of gcForest and gcForestcs are decreased by 0.58\% and 1.24\%, respectively, when general parameters are used. However, the accuracy of DBC-Forest is only reduced by 0.4\%. This phenomenon demonstrates that the binning confidence screening is more accurate than the method used by gcForestcs for threshold selection, and that instance screening is more demanding, resulting in its performance being very robust to the hyperparameter settings. 

3. Table 6 shows that the accuracies of the three models are not very different on low-dimensional datasets, and that DBC-Forest is slightly better than the other models. For example, on IMDB dataset, the accuracies of gcForest, gcForestcs and DBC-Forest are 88.81\%, 88.88\% and 89.39\%, respectively. This result indicates that binning confidence screening is more effective than the original confidence screening.

\subsection{Efficiency of binning confidence screening}

In the previous subsections, we validate the performance of our model in terms of accuracy and robustness. In this subsection, we analyze the efficiency of the binning confidence screening. Since the five datasets DIGITS, ADULT, BANK, YEAST and LETTER contain less data and are easier to predict, only two or three levels are generated in the cascade structure. Therefore, we use MNIST, EMNIST, FASHION-MNIST and IMDB for this experiment. We compare the remaining instances and accuracy at each level, and the results are shown in Figures 12 to 15.\par

\begin{figure}[H]
	\centering 
	\subfigure[Remaining instances]{ 
		\begin{minipage}{4.2cm}
			\includegraphics[width=4.2cm,height=3cm]{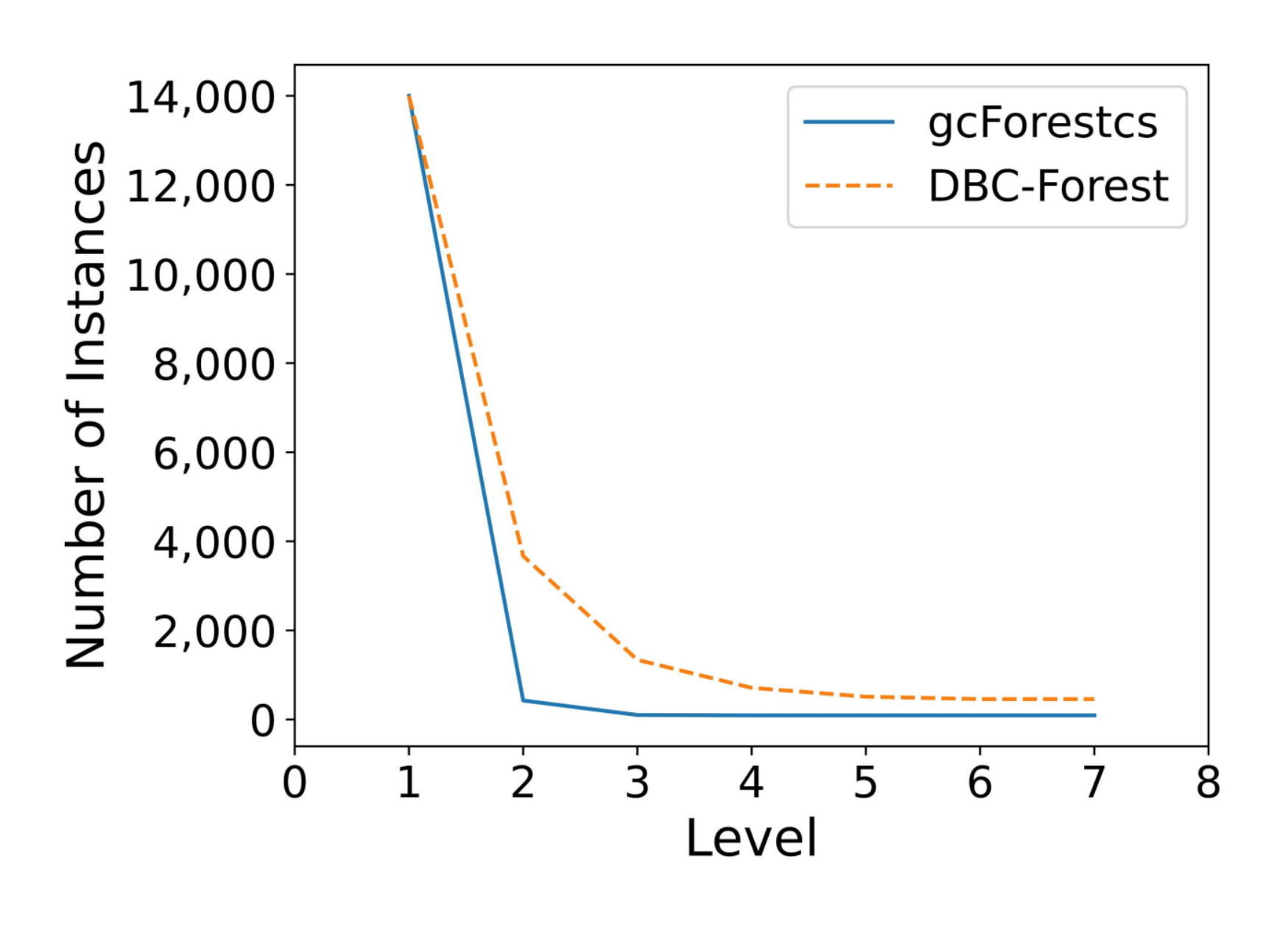}
		\end{minipage}
	}
	\subfigure[Accuracy at each level]{ 
		\begin{minipage}{4.2cm}
			\includegraphics[width=4.2cm,height=3cm]{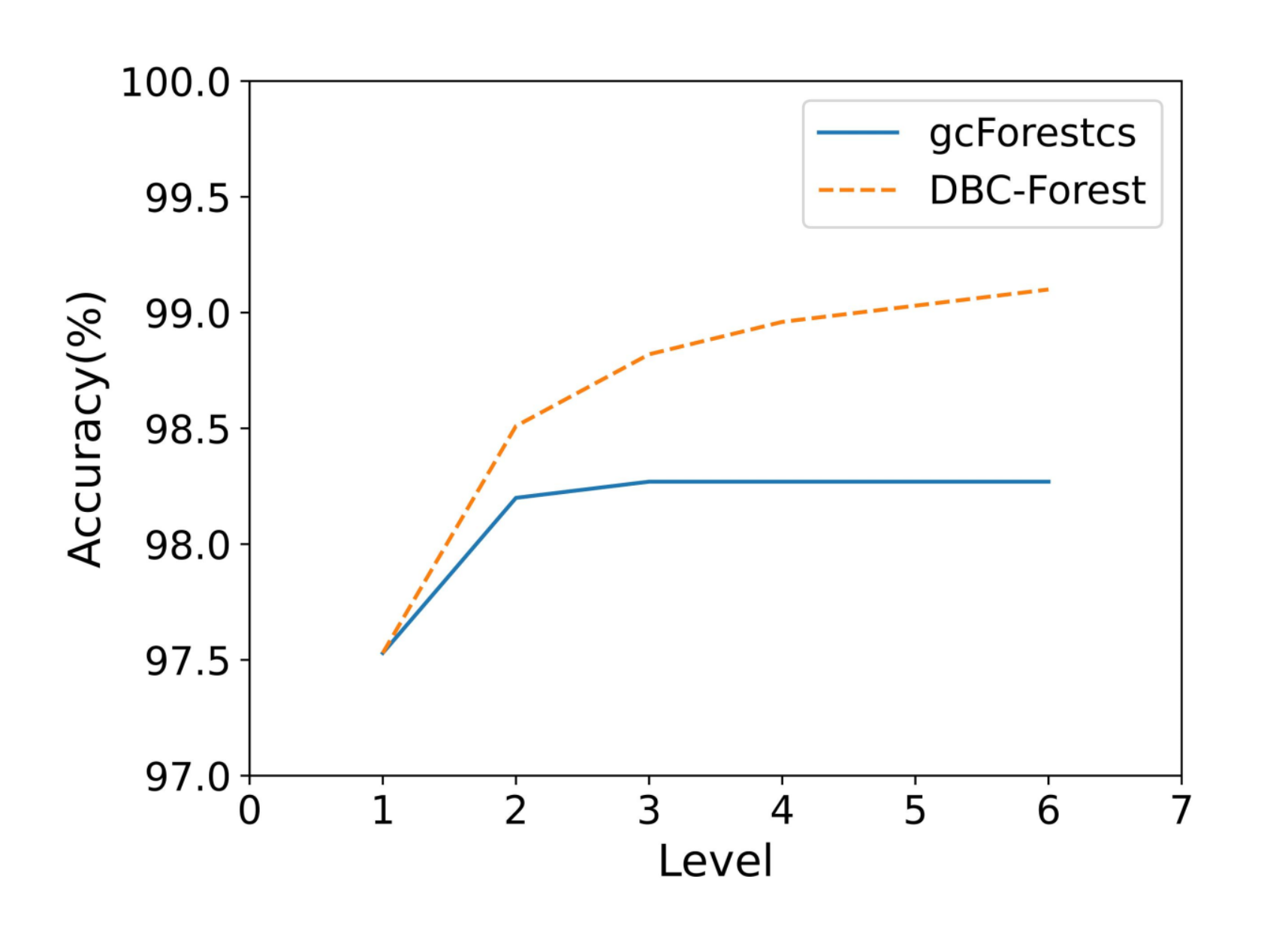} 
		\end{minipage}
	}
	\caption{Comparison of remaining instances and accuracy at each level on MNIST dataset} 
\end{figure}
\begin{figure}[h]
	\centering 
	\subfigure[Remaining instances]{ 
		\begin{minipage}{4.2cm}
			\includegraphics[width=4.2cm,height=3cm]{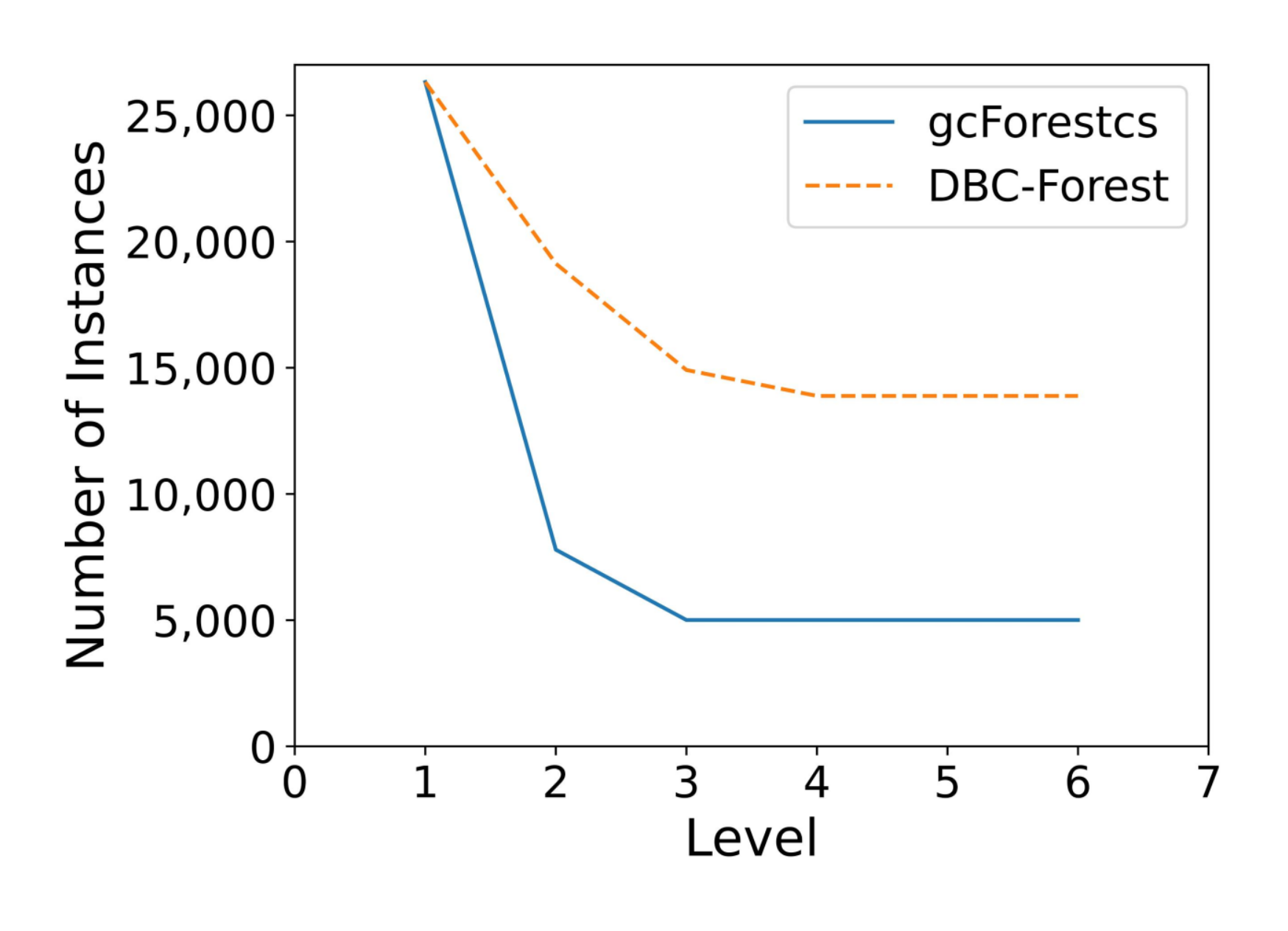} 
		\end{minipage}
	}
	\subfigure[Accuracy at each level]{ 
		\begin{minipage}{4.2cm}
			\includegraphics[width=4.2cm,height=3cm]{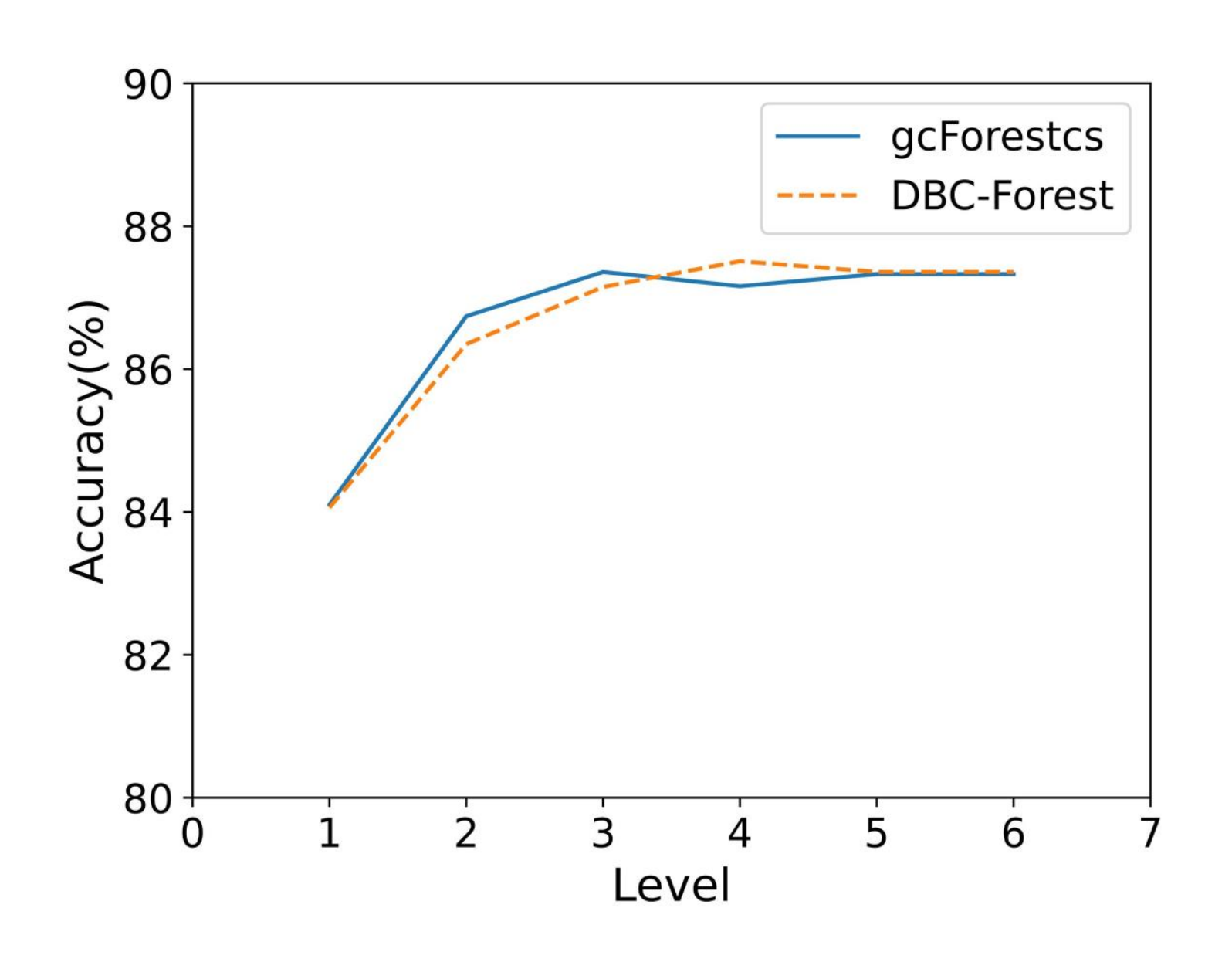} 
		\end{minipage}
	}
	\caption{Comparison of remaining instances and accuracy at each level on EMNIST dataset} 
	\subfigure[Remaining instances]{ 
		\begin{minipage}{4.2cm}
			\includegraphics[width=4.2cm,height=3cm]{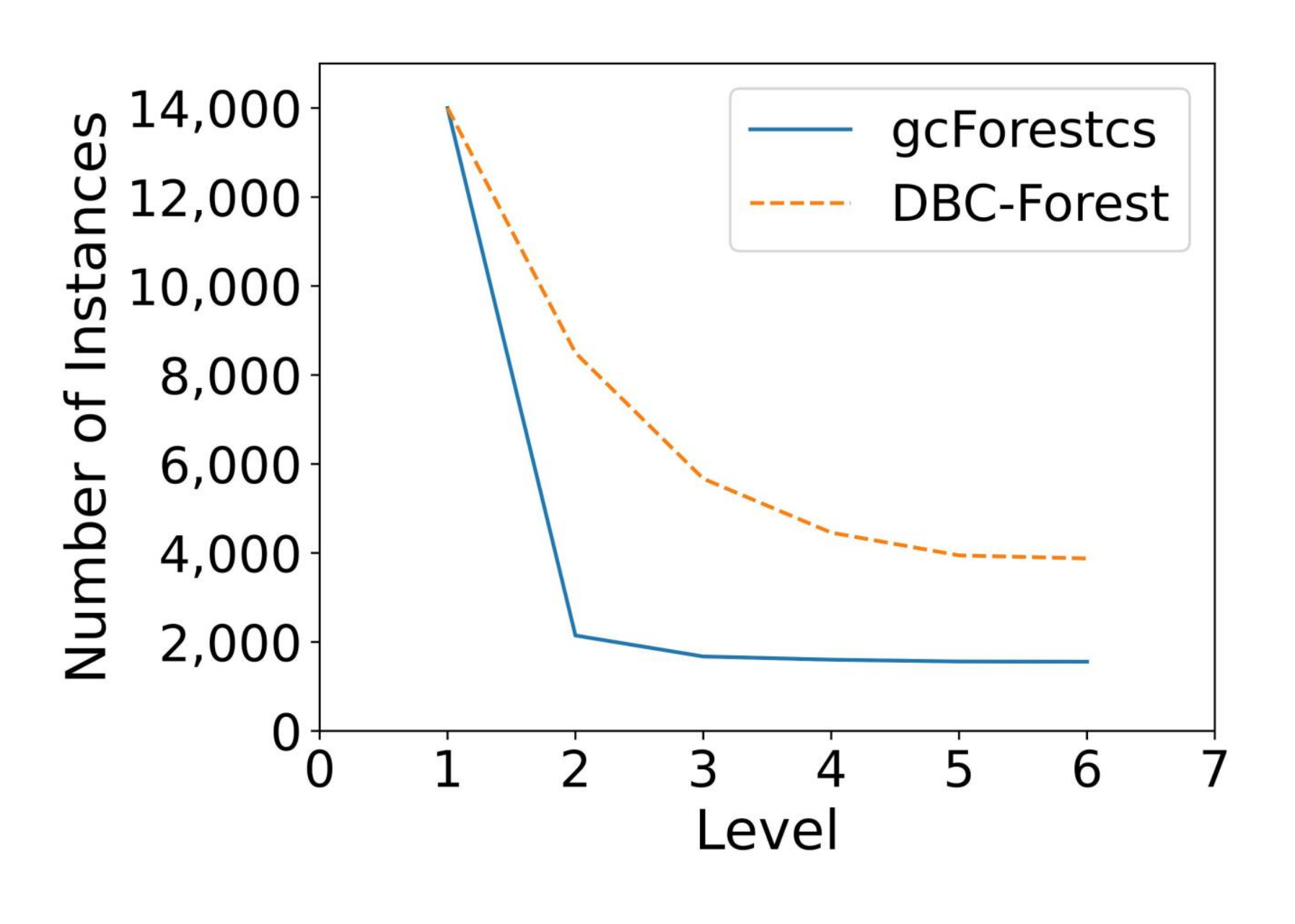} 
		\end{minipage}
	}
	\subfigure[Accuracy at each level]{ 
		\begin{minipage}{4.2cm}
			\includegraphics[width=4.2cm,height=3cm]{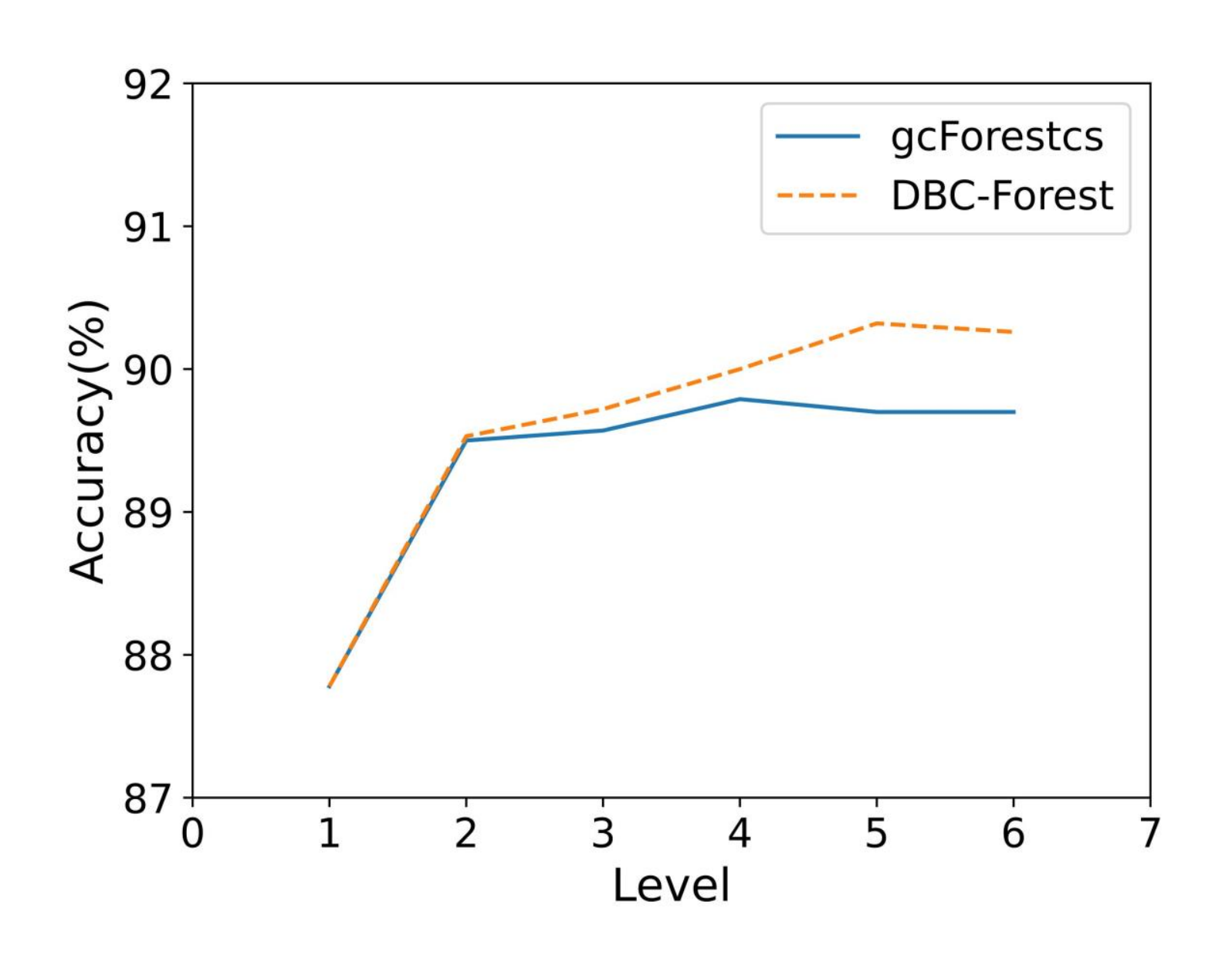} 
		\end{minipage}
	}
	\caption{Comparison of remaining instances and accuracy at each level on FASHION-MNIST dataset} 
	\centering 
	\subfigure[Remaining instances]{ 
		\begin{minipage}{4.2cm}
			\includegraphics[width=4.2cm,height=3cm]{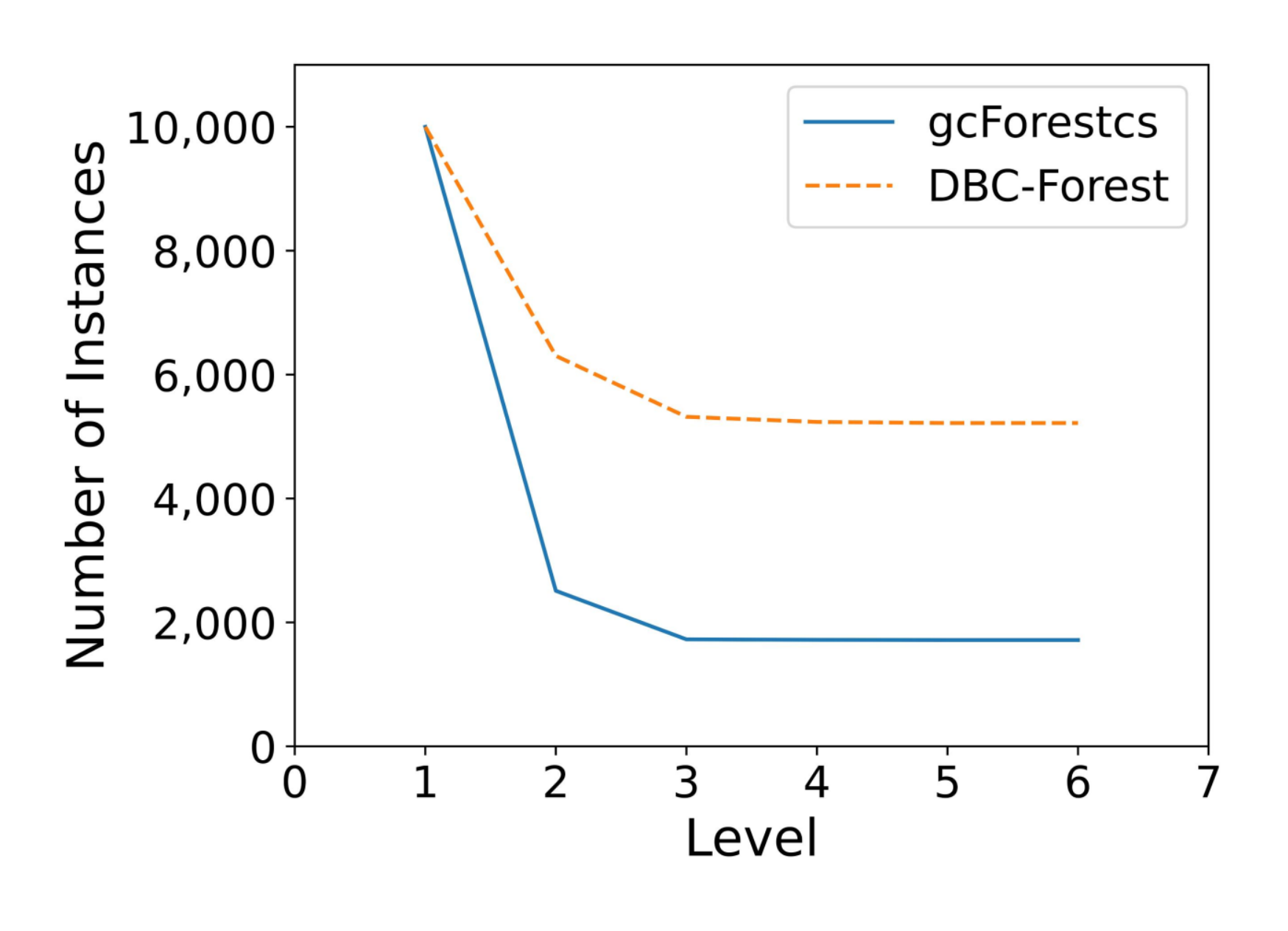} 
		\end{minipage}
	}
	\subfigure[Accuracy at each level]{ 
		\begin{minipage}{4.2cm}
			\includegraphics[width=4.2cm,height=3cm]{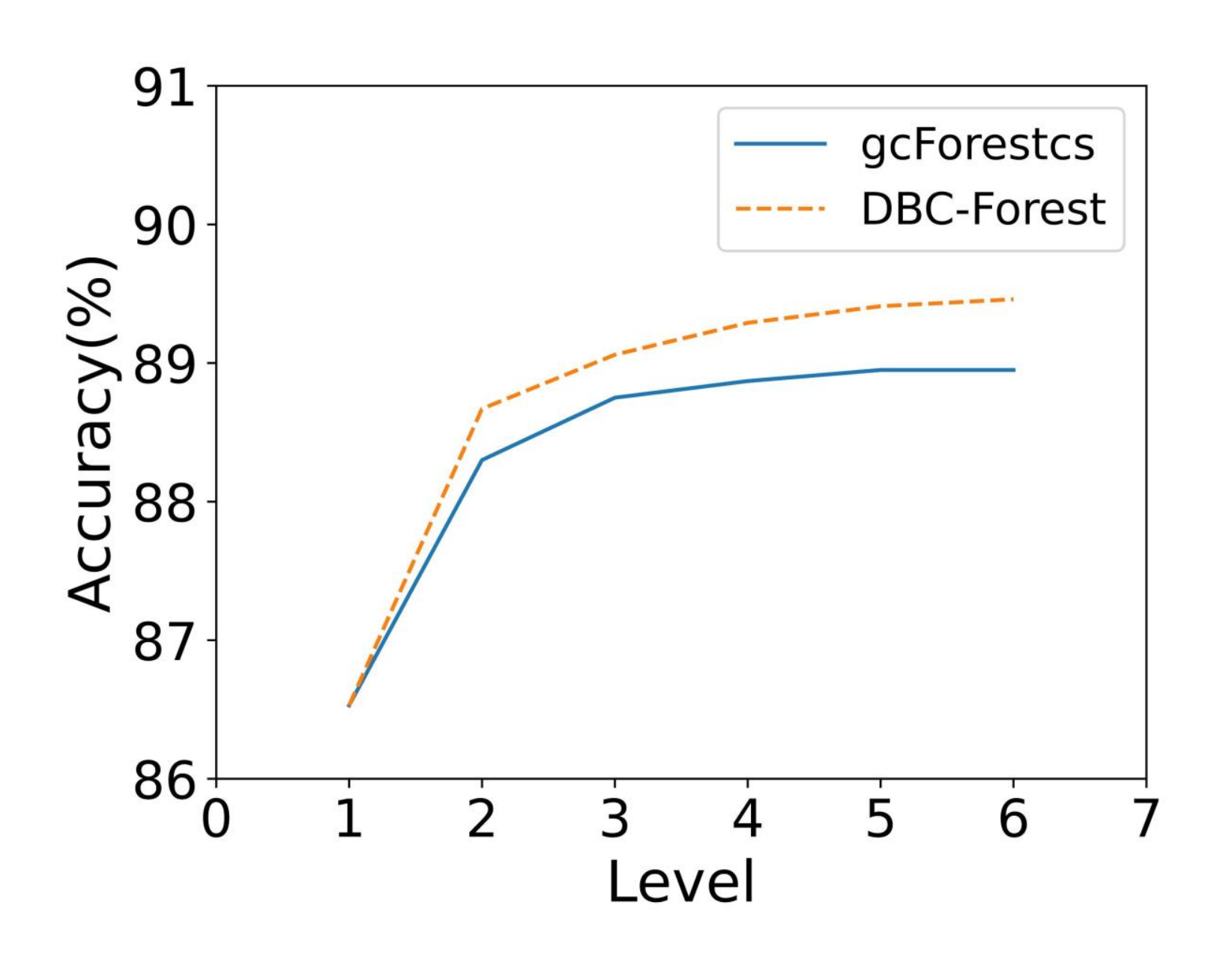} 
		\end{minipage}
	}
	\caption{Comparison of remaining instances and accuracy at each level on IMDB dataset} 
\end{figure}

Figures 12 to 15 show that although DBC-Forest retains more instances than gcForestcs, its accuracy at each level is higher than that of gcForestcs on most datasets. For example, on MNIST dataset, DBC-Forest retains about 3,700 instances, while gcForestcs retains about 400 at the second level. However, the accuracy of DBC-Forest is higher than that of gcForestcs at the second level. Almost all of the results show this phenomenon. More importantly, the curve of the remaining instances for gcForestcs drops steeply, while that of DBC-Forest drops smoothly. The reason for this is that DBC-Forest can find mis-partitioned instances, and Examples 1 and 2 illustrate this phenomenon.

\subsection{Training time}

In this subsection, to achieve  the same accuracy, we report the training time of  gcForest, gcForestcs and DBC-Forest. The accuracy and training time on nine datasets are shown in Table 7.

\begin{table*}[h]
	\centering
	\caption{{Comparison of training time with  same accuracy}}
	\setlength{\tabcolsep}{1.3mm}{
		\begin{tabular}{@{}ccccccc@{}}
			\toprule
			& \multicolumn{3}{c}{Accuracy (\%)} & \multicolumn{3}{c}{Training time (s)} \\ 
			& gcForest & gcForestcs & DBC-Forest & gcForest & gcForestcs & DBC-Forest \\  \hline
			MNIST & \textbf{98.52$\pm$0.12} & \textbf{98.52$\pm$0.11} & 98.63$\pm$0.08 & 1,798.72 & 885.27 & \textbf{437.90} \\
			DIGITS & 97.55$\pm$0.69 & 97.16$\pm$0.48 & \textbf{97.55$\pm$0.71} & 19.02 & 18.79 & \textbf{18.73} \\
			EMNIST & 86.85$\pm$0.22 & 87.24$\pm$0.23 & \textbf{87.32$\pm$0.29} & 6,848.37 & \textbf{6,382.01} & 7,515.22 \\
			FASHION-MNIST & 90.32$\pm$0.24 & 89.94$\pm$0.29 & \textbf{90.42$\pm$0.22} & 16,369.51 & 1,955.10 & \textbf{560.32} \\
			LETTER & \textbf{96.90$\pm$0.18} & \textbf{96.90$\pm$0.21} & \textbf{96.90$\pm$0.23} & 14.30 & 14.93 & \textbf{10.52} \\
			ADULT & \textbf{86.11$\pm$0.27} & \textbf{86.11$\pm$0.27} & \textbf{86.11$\pm$0.29} & 21.15 & 17.72 & \textbf{15.59} \\
			BANK & 91.48$\pm$0.22 & 91.48$\pm$0.20 & \textbf{91.50$\pm$0.18} & 20.68 & 16.82 & \textbf{15.83} \\
			YEAST & \textbf{61.45$\pm$1.87} & \textbf{61.45$\pm$2.74} & \textbf{61.45$\pm$1.37} & \textbf{8.50} & 10.32 & 9.70 \\
			IMDB & 89.10$\pm$0.08 & \textbf{89.39$\pm$0.13} & \textbf{89.39$\pm$0.32} & 8,828.60 & 802.41 & \textbf{654.32} \\ \bottomrule
	\end{tabular}}
\end{table*}
It should be noted that it is difficult to obtain exactly equal accuracy for different schemes, and the accuracies in Table 7 are therefore slightly different.

As shown in Table 7, DBC-Forest requires a shorter training time than gcForest and gcForestcs for the same or slightly higher accuracy. For example, on IMDB dataset, the training time of DBC-Forest, gcForest and gcForestcs are 654.32, 8,828.60 and 802.41 s, respectively. These results show that DBC-Forest is more effective than the comparable schemes. The reason is that the accuracy of DBC-Forest at most levels is higher than gcForestcs, as mentioned in Subsection 4.5, and hence DBC-Forest can be trained faster than the other schemes.

\section{Conclusion}

To improve the efficiency of the deep forest algorithm, gcForestcs allows some instances in high-confidence region to be passed directly to the final stage. However, there is a group of instances in this region with low prediction accuracy.  To find these instances, we propose DBC-Forest, which ranks the instances based on the confidences and puts them into bins of the same size. The accuracy of each bin is the average value of the accuracies of the instances in the bin. DBC-Forest finds a bin with an accuracy lower than the target accuracy, and uses the confidence of the last instance in the bin as the threshold. In this way, DBC-Forest effectively avoids the mis-partitioning of instances, and improves the classification performance of gcForestcs. Our experimental results demonstrate that DBC-Forest achieves better accuracy performance than other state-of-the-art models for the same parameters, and has higher robustness than gcForest and gcForestcs. We also discuss the efficiency of binning confidence screening, and show that DBC-Forest can be trained faster than the other schemes for the same accuracy.

There are many aspects to be considered in the future.

1. Although experiments show that DBC-Forest achieve best performance, the hyperparameters may not be appropriate for some models. How to select the appropriate hyperparameter for each deep forest model is worth developing.

2. binning confidence screening mechanism to avoid mis-partition instances. The hyperparameter size of bin influences DBC-Forest performance. In the future, we will design a self-adaptive screening method to replace the current mechanism.

\section*{Acknowledgement}
This work was partly supported by National Natural Science Foundation of China (61976240, 52077056), Natural Science Foundation of Hebei Province, China (Nos. F2020202013, E2020202033), and Graduate Student Innovation Program of Hebei Province (CXZZSS2021026).

\end{document}